\documentclass{article}

\usepackage[utf8]{inputenc} %
\usepackage[T1]{fontenc}    %
\usepackage{microtype}
\usepackage{graphicx}
\usepackage{booktabs} %
\usepackage[aboveskip=2pt]{subcaption} %

\usepackage[dvipsnames]{xcolor}

\usepackage{hyperref}

\usepackage[accepted]{icml2023}

\usepackage{url}            %
\usepackage{amsfonts}       %
\usepackage{nicefrac}       %
\usepackage{amsmath,amssymb}
\usepackage[aboveskip=2pt]{subcaption}
\usepackage{graphics}
\usepackage{mathtools}
\usepackage{bm}

\usepackage{siunitx}

\usepackage{tikz,pgfplots}
\usetikzlibrary{shapes,arrows,positioning,calc}
\usetikzlibrary{shapes.geometric}
\usetikzlibrary{decorations.pathmorphing}
\pgfkeys{/pgf/number format/.cd,1000 sep={}}
\newlength{\figurewidth}
\newlength{\figureheight}

\renewcommand{\paragraph}[1]{\textbf{#1}~~}

\usepackage{algpseudocode}
\usepackage{cancel}

\algrenewcomment[1]{\hfill {\color{gray}\(\triangleright\) #1}}
\algnewcommand{\LineComment}[1]{\State {\color{gray}\(\triangleright\) #1}}

\makeatletter
\newlength{\trianglerightwidth}
\algnewcommand{\LineCommentCont}[1]{\Statex \hskip\ALG@thistlm%
  \parbox[t]{\dimexpr\linewidth-\ALG@thistlm}{\hangindent=\trianglerightwidth \hangafter=1 \strut {\color{gray}$\triangleright$ #1}\strut}}
\makeatother

\usepackage[capitalize,nameinlink]{cleveref}
\crefname{section}{Sec.}{Secs.}
\crefname{appendix}{App.}{Apps.}
\crefname{algorithm}{Alg.}{Algs.}

\usepackage{xspace}

\newcommand{\ie}{\textit{i.e.}\@\xspace}
\newcommand{\cf}{\textit{cf.}\@\xspace}

\newcommand{\etal}{\textit{et~al.}\@\xspace}

\usepackage{bm}
\newcommand{\mathbold}[1]{\bm{#1}}
\newcommand{\mbf}[1]{\mathbf{#1}}

\newcommand{\MZ}{\mbf{Z}}
\newcommand{\MV}{\mbf{V}}
\newcommand{\MX}{\mbf{X}}
\newcommand{\MA}{\mbf{A}}
\newcommand{\MP}{\mbf{P}}
\newcommand{\MK}{\mbf{K}}

\newcommand{\MH}{\mbf{H}}

\newcommand{\T}{\top}
\newcommand{\vzeros}{\mbf{0}}
\newcommand{\vtheta}[0]{\mathbold{\theta}}
\newcommand{\vBeta}{\bm{B}} %

\newcommand{\vm}{\mbf{m}}
\newcommand{\vu}{\mbf{u}}
\newcommand{\vx}{\mbf{x}}
\newcommand{\vy}{\mbf{y}}

\newcommand{\GP}{\mathcal{GP}}

\newcommand{\MKzz}{\mbf{K}_{\mbf{z}\mbf{z}}}
\newcommand{\MKxx}{\mbf{K}_{\mbf{x}\mbf{x}}}
\newcommand{\MKzx}{\mbf{K}_{\mbf{z}\mbf{x}}}
\newcommand{\MKxz}{\mbf{K}_{\mbf{x}\mbf{z}}}
\newcommand{\vkzi}{\mbf{k}_{\mbf{z}i}}
\newcommand{\vkzs}{\mbf{k}_{\mbf{z}i}}
\newcommand{\vk}{\mbf{k}}

\definecolor{matplotlib-blue}{HTML}{1f77b4}
\newcommand{\N}{\mathrm{N}}

\newcommand{\comment}[1]{}

\newcommand{\vparam}{\vtheta}

\newcommand{\dkl}[3]{\mathbb{D}_{\text{KL}}^{#1}[#2 \, \|\, #3]}

\newcommand\cut[1]{}

\newcommand{\squishlist}{
   \begin{list}{$\bullet$}
    { \setlength{\itemsep}{0pt}      \setlength{\parsep}{3pt}
      \setlength{\topsep}{3pt}       \setlength{\partopsep}{0pt}
      \setlength{\leftmargin}{1.5em} \setlength{\labelwidth}{1em}
      \setlength{\labelsep}{0.5em} } }

\newcommand{\squishlisttwo}{
   \begin{list}{$\bullet$}
    { \setlength{\itemsep}{0pt}    \setlength{\parsep}{0pt}
      \setlength{\topsep}{0pt}     \setlength{\partopsep}{0pt}
      \setlength{\leftmargin}{2em} \setlength{\labelwidth}{1.5em}
      \setlength{\labelsep}{0.5em} } }

\newcommand{\squishend}{
    \end{list}  }

{}
{}
{}
{}

\newcommand{\half}{\text{$\frac{1}{2}$}}

\newcommand{\rnd}[1]{\left(#1\right)}
\newcommand{\sqr}[1]{\left[#1\right]}

\newcommand{\myexpect}{\mathbb{E}}

\newcommand{\gauss}{\text{${\cal N}$}}

\newcommand{\myvec}[1]{\ensuremath{\boldsymbol{#1}}}
\newcommand{\myvecsym}[1]{\ensuremath{\boldsymbol{#1}}}

\newcommand{\valpha}{\text{$\myvecsym{\alpha}$}}
\newcommand{\vbeta}{\text{$\myvecsym{\beta}$}}

\newcommand{\vmu}{\text{$\myvecsym{\mu}$}}

\newcommand{\vlambda}{\text{$\myvecsym{\lambda}$}}

\newcommand{\vSigma}{\text{$\myvecsym{\Sigma}$}}

\newcommand{\va}{\text{$\myvec{a}$}}
\newcommand{\vb}{\text{$\myvec{b}$}}

\newcommand{\vf}{\text{$\myvec{f}$}}

\newcommand{\vz}{\text{$\myvec{z}$}}

\newcommand{\vB}{\text{$\myvec{B}$}}

\newcommand{\vV}{\text{$\myvec{V}$}}

\newcommand{\vZ}{\text{$\myvec{Z}$}}

\newcommand{\diag}{\text{{diag}}}

\newcommand{\calD}{\text{${\cal D}$}}

\newcommand{\data}{\calD}

\newcommand{\be}{\begin{equation}}
\newcommand{\ee}{\end{equation}}
\newcommand{\bea}{\begin{eqnarray}}
\newcommand{\eea}{\end{eqnarray}}
\newcommand{\beaa}{\begin{eqnarray*}}
\newcommand{\eeaa}{\end{eqnarray*}}

\renewcommand{\mid}{\,|\,}
\newcommand{\diff}{\,\mathrm{d}}
\renewcommand{\vf}{\mbf{f}}
{}

\renewcommand{\gauss}{\mathrm{N}}

\usepackage{tabularx}
\usepackage{booktabs}

\definecolor{mycolor0}{HTML}{2C6EBA}
\definecolor{mycolor1}{HTML}{7CAC56}
\definecolor{mycolor2}{rgb}{0.8275,0.2627,0.3059}
\definecolor{mycolor3}{rgb}{0.5216,0.4392,0.7176}
\definecolor{mycolor4}{rgb}{0.8118,0.7255,0.4118}
\definecolor{mycolor5}{rgb}{0.2745,0.7176,0.8157}

\pgfplotsset{every axis/.append style={
		legend style={inner xsep=1pt, inner ysep=0.5pt, nodes={inner sep=1pt, text depth=0.1em},draw=none,fill=white}
}}    

\newcommand{\digit}[1]{\tikz[baseline=-.5ex]\node[inner sep=1pt,rounded corners=1pt,draw=black,text width=5pt,minimum width=5pt,align=center,fill=black!20]{\tiny\bf\sf#1};}

\begin{document}

\twocolumn[

\icmltitle{Memory-Based Dual Gaussian Processes for Sequential Learning}

\icmlsetsymbol{equal}{*}

\begin{icmlauthorlist}
\icmlauthor{Paul E.\ Chang}{equal,aalto}
\icmlauthor{Prakhar Verma}{equal,aalto}
\icmlauthor{S.T.\ John}{aalto,fcai}
\icmlauthor{Arno Solin}{aalto,fcai}
\icmlauthor{Mohammad Emtiyaz Khan}{riken}
\end{icmlauthorlist}

\icmlaffiliation{aalto}{Department of Computer Science, Aalto University, Finland}
\icmlaffiliation{fcai}{Finnish Center for Artificial Intelligence (FCAI)}
\icmlaffiliation{riken}{RIKEN Center for AI Project, Tokyo, Japan}

\icmlcorrespondingauthor{Paul Chang}{paul.chang@aalto.fi}

\icmlkeywords{Machine Learning, Gaussian Processes, Sequential Learning}

\vskip 0.3in
]

\printAffiliationsAndNotice{\icmlEqualContribution} %

\begin{abstract}
Sequential learning with Gaussian processes (GPs) is challenging when access to past data is limited, for example, in continual and active learning. In such cases, errors can accumulate over time due to inaccuracies in the posterior, hyperparameters, and inducing points, making accurate learning challenging. Here, we present a method to keep all such errors in check using the recently proposed dual sparse variational GP. Our method enables accurate inference for generic likelihoods and improves learning by actively building and updating a memory of past data. We demonstrate its effectiveness in several applications involving Bayesian optimization, active learning, and continual learning.
\looseness-2
\end{abstract}

\section{Introduction}
\label{sec:introduction}
Sequential decision-making requires uncertainty estimates that can be used to plan for the future. For this reason, Gaussian process (GP) models are popular for sequential problems in applications such as model-based reinforcement learning \citep{deisenroth2011pilco} and Bayesian optimization \citep{garnett_bayesoptbook_2022}.
However, exact online inference in GPs requires access to all past data, which becomes infeasible over time as the amount of data grows \citep{csato2002sparse}.
Sparse GP methods can reduce this cost, but they too assume access to all past data. For example, the popular sparse variational GP (SVGP) and variational free energy \citep[VFE,][]{titsias2009variational} methods require multiple passes through the data during the stochastic training \citep{hensman2013gaussian}. This can lead to inaccuracies for continual or active learning, where access to past data is limited and errors can accumulate over time.\looseness-1

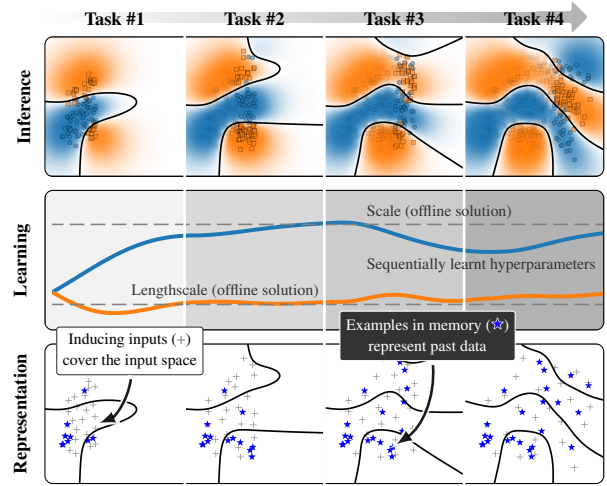
\begin{figure}[t!]
  \centering\scriptsize
  \begin{tikzpicture}

\tikzset{%
  outlinearrow/.style n args={4}{%
    -latex,
    line width=#2,
    draw=#4,
    color=#4,
    opacity=0.9,
    postaction={
      draw=#1,
      color=#1,
      line width=#2-#3,
      shorten >=2*#3,
      shorten <=#3,
    },
    outlinearrow/.default={gray}{5pt}{1pt}{black},
    outlinearrow/.initial={gray}{5pt}{1pt}{black},
  }
}

     \setlength{\figurewidth}{.9\columnwidth}
     \setlength{\figureheight}{.25\figurewidth}

	\newcommand{\representationblocks}[4]{

		\begin{scope}

			\clip[rounded corners=3pt] ($(#3)+(.125\figurewidth,-.5\figureheight)$) rectangle ++(\figurewidth,\figureheight);

			\foreach \x/\c [count=\i] in {0/05,1/15,2/20,3/30} {
				\node[fill=#2!\c,minimum width=.25\figurewidth,minimum height=\figureheight, draw, inner sep=0] (#1-\x) at ($(#3)+({\i*0.25*\figurewidth},0)$) 
				{\includegraphics[width=.25\figurewidth]{./fig/banana/streaming_banana_mem_Z_\i.png}};
			}       
			
		\end{scope}

		\draw[rounded corners=3pt,draw=#2] ($(#3)+(.125\figurewidth,-.5\figureheight)$) rectangle ++(\figurewidth,\figureheight);

		\foreach \x in {1,2,3}
		\draw[white,line width=1pt] ($(#3)+({.125\figurewidth+.25*\x*\figurewidth},-.55\figureheight)$) -- ++(0,1.1\figureheight);

		\node[rotate=90,anchor=south,outer sep=3pt] at (#1-0.west) {\textbf{#1}};
		
	} %
       
    \representationblocks{Representation}{black}{0,0}{MNIST}

	\newcommand{\learningblocks}[4]{

		\begin{scope}

			\clip[rounded corners=3pt] ($(#3)+(.125\figurewidth,-.5\figureheight)$) rectangle ++(\figurewidth,\figureheight);

			\foreach \x/\c [count=\i] in {0/05,1/15,2/20,3/30} {
				\node[fill=#2!\c,minimum width=.25\figurewidth,minimum height=\figureheight, draw, inner sep=0] (#1-\x) at ($(#3)+({\i*0.25*\figurewidth},0)$) {};
			}

			\pgfplotsset{scale only axis,hide axis}
			\node[anchor=north east] at ($(#3) + (1.15\figurewidth,.5\figureheight)$) {
\begin{tikzpicture}

\definecolor{darkgray176}{RGB}{176,176,176}
\definecolor{darkorange25512714}{RGB}{255,127,14}
\definecolor{gray}{RGB}{128,128,128}
\definecolor{steelblue31119180}{RGB}{31,119,180}

\begin{axis}[
height=\figureheight,
hide x axis,
hide y axis,
tick align=outside,
tick pos=left,
width=\figurewidth,
x grid style={darkgray176},
xmin=0, xmax=80,
xtick style={color=black},
y grid style={darkgray176},
ymin=0.3, ymax=2.5,
ytick style={color=black}
]
\addplot [line width=1.5pt, steelblue31119180]
table {%
0 1
1 1.06436415512332
2 1.13043693597578
3 1.19765382658596
4 1.26536545430928
5 1.33283991258984
6 1.39927154011482
7 1.46380911169103
8 1.52560560681904
9 1.58387717052038
10 1.63795302019072
11 1.68730494083199
12 1.73155534233478
13 1.77046902945529
14 1.80393559937112
15 1.83194875071352
16 1.85458712879618
17 1.8719992994435
18 1.88439341671257
19 1.89203049804248
20 1.89521932645407
21 1.89521932645407
22 1.90068069044827
23 1.90834387349468
24 1.91802987701796
25 1.92945613307663
26 1.9422566801079
27 1.95601307435945
28 1.97028978388288
29 1.98466816757032
30 1.99877464708262
31 2.01230065699207
32 2.02501384415828
33 2.03676134345133
34 2.04746658723372
35 2.05712110678794
36 2.06577245025165
37 2.07350901519193
38 2.08044254362411
39 2.08668937185297
40 2.09235217349132
41 2.09750455424565
42 2.09750455424565
43 2.09137280888056
44 2.07567570295652
45 2.05231538570502
46 2.02323587518456
47 1.99030749417266
48 1.95522360202135
49 1.91941932855608
50 1.88402717305671
51 1.84987075049279
52 1.81748963257851
53 1.78719123160284
54 1.75912649108676
55 1.73337934010285
56 1.71005171227285
57 1.68932422827403
58 1.67147944231443
59 1.65688519160358
60 1.64594429513257
61 1.6390222327016
62 1.63636808039392
63 1.63636808039392
64 1.63723777284711
65 1.6425080304638
66 1.65276047045441
67 1.66811859089182
68 1.68819102724048
69 1.71211515523872
70 1.7386749817636
71 1.76646819703721
72 1.79410175745617
73 1.82038880395051
74 1.84450790505057
75 1.86608588480719
76 1.8851848273448
77 1.90220132563597
78 1.91770723337173
79 1.9322713491474
80 1.94630241408922
81 1.95994645829322
82 1.97305678676674
83 1.98523724135815
};
\addplot [line width=1.5pt, darkorange25512714]
table {%
0 1
1 0.93796051742217
2 0.879536445294823
3 0.825848610078665
4 0.778216978021431
5 0.738122822834278
6 0.70696834911253
7 0.685604974968567
8 0.673947917992055
9 0.671056263321403
10 0.675557675042729
11 0.686018491870229
12 0.701096392184188
13 0.719540668663435
14 0.740137056733026
15 0.761656625161557
16 0.782839671747918
17 0.802430543849707
18 0.819264485615413
19 0.832386401532854
20 0.841161884241083
21 0.841161884241083
22 0.847604979924385
23 0.851227250393347
24 0.851911157621675
25 0.849907696140059
26 0.84576854800382
27 0.840241192586672
28 0.83415492186113
29 0.828315665871571
30 0.823419583512872
31 0.819991265425523
32 0.818349044511874
33 0.818595804227049
34 0.82063015445257
35 0.824171826897688
36 0.828796951104667
37 0.83398197206752
38 0.839157003680091
39 0.843768428519666
40 0.847346154571597
41 0.849565176805707
42 0.849565176805707
43 0.861523469149375
44 0.880459451311683
45 0.903156957899187
46 0.92594973271477
47 0.944989232749982
48 0.957032727192849
49 0.960361754461877
50 0.955106539073994
51 0.942864150803367
52 0.926087268450151
53 0.907576659087081
54 0.890083800337087
55 0.875942666363251
56 0.866745320305522
57 0.863164580231431
58 0.86498010094372
59 0.871246462189389
60 0.880504813832233
61 0.890998684356796
62 0.90091404187708
63 0.90091404187708
64 0.91242956128208
65 0.923611131143415
66 0.932786329960889
67 0.938858462722245
68 0.941505324448634
69 0.941186035464769
70 0.938969314093674
71 0.936251856060089
72 0.934438389867602
73 0.934646075714177
74 0.937490691142081
75 0.942992406559882
76 0.950603795299
77 0.959338462743878
78 0.967976690201558
79 0.975323192710048
80 0.980471233747844
81 0.983002523298245
82 0.983059555377323
83 0.981272658422181
};
\path [draw=gray, semithick, dash pattern=on 5.55pt off 2.4pt]
(axis cs:-0.5,0.807844163463287)
--(axis cs:100,0.807844163463287);

\path [draw=gray, semithick, dash pattern=on 5.55pt off 2.4pt]
(axis cs:-0.5,2.06925298416419)
--(axis cs:100,2.06925298416419);

\end{axis}

\end{tikzpicture}};
			
		\end{scope}

		\draw[rounded corners=3pt,draw=#2] ($(#3)+(.125\figurewidth,-.5\figureheight)$) rectangle ++(\figurewidth,\figureheight);

		\foreach \x in {1,2,3}
		\draw[white,line width=1pt] ($(#3)+({.125\figurewidth+.25*\x*\figurewidth},-.55\figureheight)$) -- ++(0,1.1\figureheight);

		\node[rotate=90,anchor=south,outer sep=3pt] at (#1-0.west) {\textbf{#1}};
		
	} %

		\begin{scope}

			\clip[rounded corners=3pt] ($(0,1.1\figureheight)+(.125\figurewidth,-.5\figureheight)$) rectangle ++(\figurewidth,\figureheight);

			\foreach \x/\c [count=\i] in {0/05,1/15,2/20,3/30} {
				\node[fill=black!\c,minimum width=.25\figurewidth,minimum height=\figureheight, draw, inner sep=0] (Learning-\x) at ($(0,1.1\figureheight)+({\i*0.25*\figurewidth},0)$) {};
			}

			\pgfplotsset{scale only axis,hide axis}
			\node[anchor=north east] at ($(0,1.1\figureheight) + (1.15\figurewidth,.5\figureheight)$) {
\begin{tikzpicture}

\definecolor{darkgray176}{RGB}{176,176,176}
\definecolor{darkorange25512714}{RGB}{255,127,14}
\definecolor{gray}{RGB}{128,128,128}
\definecolor{steelblue31119180}{RGB}{31,119,180}

\begin{axis}[
height=\figureheight,
hide x axis,
hide y axis,
tick align=outside,
tick pos=left,
width=\figurewidth,
x grid style={darkgray176},
xmin=0, xmax=80,
xtick style={color=black},
y grid style={darkgray176},
ymin=0.3, ymax=2.5,
ytick style={color=black}
]
\addplot [line width=1.5pt, steelblue31119180]
table {%
0 1
1 1.06436415512332
2 1.13043693597578
3 1.19765382658596
4 1.26536545430928
5 1.33283991258984
6 1.39927154011482
7 1.46380911169103
8 1.52560560681904
9 1.58387717052038
10 1.63795302019072
11 1.68730494083199
12 1.73155534233478
13 1.77046902945529
14 1.80393559937112
15 1.83194875071352
16 1.85458712879618
17 1.8719992994435
18 1.88439341671257
19 1.89203049804248
20 1.89521932645407
21 1.89521932645407
22 1.90068069044827
23 1.90834387349468
24 1.91802987701796
25 1.92945613307663
26 1.9422566801079
27 1.95601307435945
28 1.97028978388288
29 1.98466816757032
30 1.99877464708262
31 2.01230065699207
32 2.02501384415828
33 2.03676134345133
34 2.04746658723372
35 2.05712110678794
36 2.06577245025165
37 2.07350901519193
38 2.08044254362411
39 2.08668937185297
40 2.09235217349132
41 2.09750455424565
42 2.09750455424565
43 2.09137280888056
44 2.07567570295652
45 2.05231538570502
46 2.02323587518456
47 1.99030749417266
48 1.95522360202135
49 1.91941932855608
50 1.88402717305671
51 1.84987075049279
52 1.81748963257851
53 1.78719123160284
54 1.75912649108676
55 1.73337934010285
56 1.71005171227285
57 1.68932422827403
58 1.67147944231443
59 1.65688519160358
60 1.64594429513257
61 1.6390222327016
62 1.63636808039392
63 1.63636808039392
64 1.63723777284711
65 1.6425080304638
66 1.65276047045441
67 1.66811859089182
68 1.68819102724048
69 1.71211515523872
70 1.7386749817636
71 1.76646819703721
72 1.79410175745617
73 1.82038880395051
74 1.84450790505057
75 1.86608588480719
76 1.8851848273448
77 1.90220132563597
78 1.91770723337173
79 1.9322713491474
80 1.94630241408922
81 1.95994645829322
82 1.97305678676674
83 1.98523724135815
};
\addplot [line width=1.5pt, darkorange25512714]
table {%
0 1
1 0.93796051742217
2 0.879536445294823
3 0.825848610078665
4 0.778216978021431
5 0.738122822834278
6 0.70696834911253
7 0.685604974968567
8 0.673947917992055
9 0.671056263321403
10 0.675557675042729
11 0.686018491870229
12 0.701096392184188
13 0.719540668663435
14 0.740137056733026
15 0.761656625161557
16 0.782839671747918
17 0.802430543849707
18 0.819264485615413
19 0.832386401532854
20 0.841161884241083
21 0.841161884241083
22 0.847604979924385
23 0.851227250393347
24 0.851911157621675
25 0.849907696140059
26 0.84576854800382
27 0.840241192586672
28 0.83415492186113
29 0.828315665871571
30 0.823419583512872
31 0.819991265425523
32 0.818349044511874
33 0.818595804227049
34 0.82063015445257
35 0.824171826897688
36 0.828796951104667
37 0.83398197206752
38 0.839157003680091
39 0.843768428519666
40 0.847346154571597
41 0.849565176805707
42 0.849565176805707
43 0.861523469149375
44 0.880459451311683
45 0.903156957899187
46 0.92594973271477
47 0.944989232749982
48 0.957032727192849
49 0.960361754461877
50 0.955106539073994
51 0.942864150803367
52 0.926087268450151
53 0.907576659087081
54 0.890083800337087
55 0.875942666363251
56 0.866745320305522
57 0.863164580231431
58 0.86498010094372
59 0.871246462189389
60 0.880504813832233
61 0.890998684356796
62 0.90091404187708
63 0.90091404187708
64 0.91242956128208
65 0.923611131143415
66 0.932786329960889
67 0.938858462722245
68 0.941505324448634
69 0.941186035464769
70 0.938969314093674
71 0.936251856060089
72 0.934438389867602
73 0.934646075714177
74 0.937490691142081
75 0.942992406559882
76 0.950603795299
77 0.959338462743878
78 0.967976690201558
79 0.975323192710048
80 0.980471233747844
81 0.983002523298245
82 0.983059555377323
83 0.981272658422181
};
\path [draw=gray, semithick, dash pattern=on 5.55pt off 2.4pt]
(axis cs:-0.5,0.807844163463287)
--(axis cs:100,0.807844163463287);

\path [draw=gray, semithick, dash pattern=on 5.55pt off 2.4pt]
(axis cs:-0.5,2.06925298416419)
--(axis cs:100,2.06925298416419);

\end{axis}

\end{tikzpicture}};
			
		\end{scope}

		\draw[rounded corners=3pt,draw=black] ($(0,1.1\figureheight)+(.125\figurewidth,-.5\figureheight)$) rectangle ++(\figurewidth,\figureheight);

		\foreach \x in {1,2,3}
		\draw[white,line width=1pt] ($(0,1.1\figureheight)+({.125\figurewidth+.25*\x*\figurewidth},-.55\figureheight)$) -- ++(0,1.1\figureheight);

		\node[rotate=90,anchor=south,outer sep=3pt] at (Learning-0.west) {\textbf{Learning}};

	\newcommand{\inferenceblocks}[4]{

		\begin{scope}

			\clip[rounded corners=3pt] ($(#3)+(.125\figurewidth,-.5\figureheight)$) rectangle ++(\figurewidth,\figureheight);

			\foreach \x/\c [count=\i] in {0/05,1/15,2/20,3/30} {
				\node[fill=#2!\c,minimum width=.25\figurewidth,minimum height=\figureheight, draw, inner sep=0] (#1-\x) at ($(#3)+({\i*0.25*\figurewidth},0)$) 
				{\includegraphics[width=.25\figurewidth]{./fig/banana/streaming_banana_tsvgp_\i_g.png}};
			}       
			
		\end{scope}

		\draw[rounded corners=3pt,draw=#2] ($(#3)+(.125\figurewidth,-.5\figureheight)$) rectangle ++(\figurewidth,\figureheight);

		\foreach \x in {1,2,3}
		\draw[white,line width=1pt] ($(#3)+({.125\figurewidth+.25*\x*\figurewidth},-.55\figureheight)$) -- ++(0,1.1\figureheight);

		\node[rotate=90,anchor=south,outer sep=3pt] at (#1-0.west) {\textbf{#1}};
		
	} %

	\inferenceblocks{Inference}{black}{0,2.2\figureheight}{} 

    \tikzstyle{myarrow} = [draw=none, single arrow, minimum height=7mm, minimum width=2pt, single arrow head extend=4pt, fill=black!30, anchor=center, rotate=0, inner sep=2pt, rounded corners=1pt]  
    \node[myarrow, anchor=south west] at ($(Inference-3.north)+(1em,.7em)$) {};
    \node[rectangle, anchor=south west, right color=black!30, left color=white, minimum width=.9\figurewidth, minimum height=2pt,inner sep=2pt] at ($(Inference-0.north west)+(0,.7em)$) {};

    \foreach \x [count=\i from 0] in {1,2,3,4} 
      \node[anchor=south,outer sep=2pt] at (Inference-\i.north) {\bf Task \#\x};

    \definecolor{steelblue31119180}{RGB}{31,119,180}
    \node at ($(Learning-2.north) + (.6cm,-.3cm)$) {\tiny \textcolor{black!80}{Scale (offline solution)}};
    \node at ($(Learning-3.north) + (-.7cm,-1.0cm)$) {\tiny \textcolor{black!80}{Sequentially learnt hyperparameters}};	
    \node at ($(Learning-2.north) + (-2.25cm,-1.35cm)$) {\tiny \textcolor{black!80}{Lengthscale (offline solution)}};    	
     
     \node[align=center,fill=white,draw=black!80,rounded corners=1pt] (rlab1) at ($(Representation-0.north) + (.2cm,-.1cm)$) {\tiny Inducing inputs (\textcolor{gray}{+}) \\ \tiny cover the input space};
     \node[align=center,draw=white,fill=black!80,rounded corners=1pt] (rlab2) at ($(Representation-1.north) + (2.3cm,.1cm)$) {\tiny\textcolor{white}{Examples in memory (}\includegraphics[width=.8em]{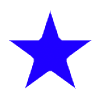}\tiny\textcolor{white}{)}\\ \textcolor{white}{\tiny represent past data}};
     
     \draw[outlinearrow={black}{2.5pt}{1.5pt}{white}] (rlab1) to[bend left=30] ++(-.5cm,-1cm);
     \draw[outlinearrow={black}{2.5pt}{1.5pt}{white}] (rlab2) to[bend left=30] ++(-.5cm,-1.5cm);
     
  \end{tikzpicture}
  \vspace*{-10pt}
  \newcommand{\mystar}{\protect\tikz[baseline=-.5ex]\protect\node[star, star point ratio=.5, fill=blue,inner sep=2pt,rotate=180]{};}
  \caption{Our sequential learning method provides an accurate posterior (top), kernel hyperparameters (middle), and sparse representation (bottom). A key contribution is to add memory of relevant past data (shown in the bottom row with \mystar) in addition to inducing inputs (shown with +).} %
  \label{fig:teaser}
  \vspace*{-1em}
\end{figure}

Errors can arise from multiple sources:~inaccurate posteriors, wrong hyperparameter values, or poor sparse representations. Various techniques can be used to control such errors, but past attempts struggled to find a coherent solution to the problem.
Initially, \citet{csato2002sparse} used expectation propagation (EP) for inference and a projection method for sparsity but did not estimate hyperparameters.
More recently, \citet{bui2017streaming} did estimate hyperparameters but found EP to perform worse than variational inference. \citet{maddox2021conditioning} did not optimize hyperparameters or use the evidence lower bound (ELBO) objective but instead resorted to a Laplace approximation for non-Gaussian likelihoods. Their method to obtain sparse representation uses a gradient method combined with pivoted Cholesky. \citet{kapoor2021variational} attempt to improve performance by using a structured
covariance during inference, but the cost grows with tasks which is due to the increasing size of the sparse representation. These attempts use a mix of methods to control various errors, which all lead to complications. We aim to build a single coherent method by tackling various errors simultaneously; see \cref{fig:teaser} for an overview. \looseness-1

We adapt the dual-SVGP method of \citet{adam2021dual} to perform sequential learning with generic likelihoods.
A key contribution of our work is to improve performance by adding a memory set of past examples. A sufficiently large memory can achieve arbitrarily close performance to the batch case, and the performance loss can be minimized by choosing the set carefully. %
For such selection, we propose a new score called the \emph{Bayesian leverage score} which uses the dual parameters of the dual-SVGP posterior to characterize relative importance of past examples.
Overall, our memory-based SVGP enables us to tackle the various errors arising in sequential learning. It also avoids the complications arising in previous work on sequential learning, as discussed below:   
\begin{enumerate}
  \item Unlike \citet{csato2002sparse}, we aim to minimize the ELBO which aligns better with \citet{bui2017streaming} who also found the ELBO to give better results than EP.
Our dual parameters can be seen as an estimate of those used in \citet[Lemma~1]{csato2002sparse}.
  \item Similarly to \citet{bui2017streaming}, we obtain a pseudo-data interpretation but it is derived from the dual parameterization. Our approach directly addresses the issues they aim to solve and is more straightforward.%
  \item Unlike \citet{maddox2021conditioning}, we estimate hyperparameters and use a variational method for inference.
  \item Unlike \citet{kapoor2021variational}, we use a fixed number of inducing inputs and instead increase the memory size. The cost grows linearly as opposed to a cubically increasing cost of their method. We also do not need any regularization for hyperparameters.
\end{enumerate}

Our use of memory is similar to recent continual learning methods \citep{nguyen2017variational, titsias2019functional, pan2020continual, khan2021knowledge}, but we select the memory via an extension of leverage scores \citep{cook1980characterizations, alaoui2015fast}. We demonstrate effectiveness of our approach on several applications involving Bayesian optimization, active learning, and continual learning.

\section{Sequential Learning with a GP}
We consider sequential learning in models with a GP prior over functions, $f \sim \GP(0, \kappa_{\vtheta})$. The prior is characterized by a covariance (kernel) function $\kappa_{\vtheta}(\vx,\vx')$, where $\vx, \vx'$ are input vectors and $\vtheta$ denotes the hyperparameters. Observations $(\vx_i, y_i)$ are modeled by the likelihood $p(y_i \mid f_i)$
given a function value $f_i = f(\vx_i)$.
In the sequential setting, we first compute the posterior on $\mathcal{D}_\textrm{old} = (\MX_{\textrm{old}},\vy_{\textrm{old}})$ where $\MX_{\text{old}}$ is a matrix containing all the past inputs $\vx_i^\top$ as rows and $\vy_{\text{old}}$ is a vector containing all the past outputs $y_i$. Then, when new data $\mathcal{D}_\textrm{new}= (\MX_{\textrm{new}},\vy_{\textrm{new}})$ is observed, our goal is to update the posterior and hyperparameters.

The posterior inference in the sequential setting is challenging because the cost grows cubically in the size of data (denoted by $n$).
We can see this by expressing the posterior $p(f_i|\vy)$ as follows \citep[Lemma~1]{csato2002sparse}, 
\begin{align}
  \myexpect_{p(f_i\mid\vy)}[f_i] &= \vk_{\vx i}^\top \valpha , \label{eq:gp_pred}\\ 
  \textrm{Var}_{p(f_i\mid\vy)}[f_i] &= \kappa_{ii} - \vk_{\vx i}^\top ( \MKxx + \diag(\vbeta)^{-1})^{-1} \vk_{\vx i}, \nonumber
\end{align}
where $\valpha$ and $\vbeta$ are vectors of $n$ dual parameters,
\begin{equation}
\begin{split} \label{eq:dual_param}
\alpha_i &=\myexpect_{p(f_i \mid \vy)}[\nabla_{f_i}\log p(y_i \mid f_i)], \\
\beta_i &= \myexpect_{p(f_i \mid \vy)}[-\nabla^2_{f_i}\log p(y_i \mid f_i)] ,
\end{split}
\end{equation}
respectively; a derivation is given in \cref{appendix:opper_equiv}.
Here, we use $\MKxx$ to denote the $n \times n$ matrix with $\kappa(\vx_i,\vx_j)$ as the $ij$\textsuperscript{th} entry. %
Similarly, $\vk_{\vx i}$ denotes a vector where each $j$\textsuperscript{th} element is $\kappa(\vx_i, \vx_j)$, and $\kappa_{ii} = \kappa(\vx_i, \vx_i)$.
The parameters $(\alpha_i, \beta_i)$ are the dual parameters that arise in the dual formulations used in, for example, support vector machines \citep{cortes1995support}, whose origins are found in work by \citet{kimeldorf1971some}.
The posterior holds for generic likelihoods, even a non-Gaussian one, and no approximations are involved.

More importantly, the expressions in \cref{eq:gp_pred} and \cref{eq:dual_param} clearly show the challenge of sequential learning, where  as data grows, so do $(\valpha,\vbeta)$, and na\"ive computation of the posterior is now $O((n_{\textrm{old}}+n_{\textrm{new}})^3)$ due to the matrix inversion, quickly making inference infeasible. \citet{csato2002sparse} suggested storing and using only a subset of past data to reduce the computation. They estimate $(\alpha_i,\beta_i)$ using a Gaussian approximation to $p(f_i\mid \vy)$, obtained by EP, but did not consider updating the hyperparameters.

In the full-batch case, scaling can be improved using the SVGP method \citep{titsias2009variational, hensman2013gaussian}, which optimizes an ELBO to select hyperparameters $\vtheta$, inducing inputs $\MZ = (\vz_1, \ldots,\vz_m)$ with $m\ll n$, and the posterior $q_{\vu}(\vu) = \mathrm{N}(\vu\mid\vm,\MV)$ defined over functions $\vu_i = f(\vz_i)$ at $\vz_i$. The ELBO is given by
\begin{equation}\label{eq:elbo_svgp_offline}
  \mathcal{L}_{\text{batch}} = \sum_{\mathclap{i \in \mathcal{D}_{\textrm{old}} \cup \mathcal{D}_{\textrm{new}}}} \mathbb{E}_{q_{\vu}(f_i)} [\log p(y_i \mid f_i)]- \dkl{}{q_{\vu}(\vu)}{p_{\vtheta}{(\vu)}},
\end{equation} 
which uses the following posterior predictive distribution,
\begin{equation}
  q_{\vu}(f_i) = \mathrm{N}(f_i \mid \va_i^\top\vm, \kappa_{ii} - \va_i^\top \MKzz \va_i + \va_i^\top \MV \va_i), \label{eq:prediction}
\end{equation}
where $\va_i^\top = \vkzi^\top \MKzz^{-1}$, and $p_{\vtheta}(\vu)$ is the GP prior over $\vu$. The last term in the bound is the Kullback--Leibler divergence (KLD). The method assumes that all data is available throughout training, so it does not directly apply in the sequential setting. %

\citet{bui2017streaming} extend the SVGP method in an ad-hoc way by adding two additional KLD terms to match the previous posterior and prior at the old inducing inputs.
Let $\MZ_{\text{old}}$ denote the old inducing inputs with $\vu_{\text{old}} = f(\MZ_{\text{old}})$ and posterior predictive $q_{\vu_{\text{old}}}(f_i)$. Also, let $p_{\vtheta_{\text{old}}}(f_i)$ denote the prior for the old hyperparameters $\vtheta_{\text{old}}$.
Given the new data $\data_{\text{new}}$, \citeauthor{bui2017streaming} modify the ELBO as follows, 
\begin{align}
   &\textstyle\sum_{i \in \mathcal{D}_{\text{new}}} \mathbb{E}_{q_{\vu}(f_i)}  [\log p(y_i \mid f_i)]
   - \dkl{}{q_{\vu}(\vu)}{p_{\vtheta}{(\vu)}}  \nonumber \\
   &\qquad\qquad+\dkl{}{q_{\vu}(\vu_{\textrm{old}})}{p_{\vtheta_{\textrm{old}}}(\vu_{\textrm{old}}))} \nonumber \\ 
   &\qquad\qquad- \dkl{}{q_{\vu}(\vu_{\textrm{old}})}{q_{\vu_{\textrm{old}}}(\vu_{\textrm{old}})} ,
 \label{eq:elbo_bui}
\end{align}
where the last two KLD terms are the newly added terms which are meant to penalize deviations from both the old prior and posterior. This is an indirect way to keep the solutions close to the full-batch case, but it does not always work well (we give empirical evidence in \cref{sec:experiments}).

\citet{maddox2021conditioning} revisit the approach of \citet{bui2017streaming} in the context of Bayesian optimization and active learning. They simplify the derivation for the Gaussian case via pseudo-data. They resort to a Laplace approximation to handle non-Gaussian likelihoods. Laplace approximation is a more `local' one than those obtained by variational methods and it can give worse results \citep{opper2009variational}. In addition, they do not optimize hyperparameters. %

We will now show that the above difficulties can be alleviated by using the method of \citet{adam2021dual}: we can estimate the parameterization of \citet{csato2002sparse}, improve the ELBO of \citet{bui2017streaming}, improve over Laplace approximation of \citet{maddox2021conditioning}, and, unlike~\citet{kapoor2021variational}, keep the computational cost from exploding.   \looseness-1

\section{A Memory-based Dual-SVGP Method}
\label{sec:methods}
We adapt the dual-SVGP method of \citet{adam2021dual} to enable accurate inference in the sequential case. We first describe the dual form of the SVGP solution and then use it to derive a new memory-based objective to perform sequential learning. Then, we optimize the new objective by using the dual-SVGP algorithm to update the posterior, hyperparameters, and inducing inputs. Finally, we present the new Bayesian leverage score to select the memory.

\subsection{The Dual Form of the SVGP Solution}
The dual form of the sequential SVGP solution has a strikingly similar form to the full-GP case given in \cref{eq:gp_pred}. Consider the following ELBO defined over the old data,
\begin{equation}
  \mathcal{L}_{\text{old}} = \sum_{\mathclap{i \in \mathcal{D}_{\textrm{old}} }} \mathbb{E}_{q_{\vu}(f_i)} [\log p(y_i \mid f_i)]- \dkl{}{q_{\vu}(\vu)}{p_{\vtheta}{(\vu)}}.
   \label{eq:elbo_old}
\end{equation} 
\citet{adam2021dual} show that a stationary point $q_{\vu}^\text{old}(\vu) = \N(\vu \mid \vm^\text{old}, \MV^\text{old})$ has a dual form for its natural parameters:
\begin{align}
   (\MV^{\text{old}})^{-1} \vm^{\text{old}} &= \textstyle \sum_{i\in\data_{\textrm{old}}} \va_i \hat{\beta}_i^\text{old} \hat{y}_i^\text{old} ,\label{eq:nat1} \\
   (\MV^{\text{old}})^{-1} &= \textstyle \sum_{i\in\data_{\textrm{old}}} \va_i \hat{\beta}_i^{\text{old}} \va_i^\top + \MKzz^{-1} , \label{eq:nat2}
\end{align}
where $\hat{y}_i^\text{old} = \hat{\alpha}_i^{\text{old}}/\hat{\beta}_i^{\text{old}} +  \va_i^\top \vm^{\text{old}}$ is a `pseudo' output, and  $(\hat{\alpha}_i^\text{old}, \hat{\beta}_i^\text{old})$ are defined as
\begin{equation}
\begin{aligned}
   \hat{\alpha}_i^\text{old} &= \myexpect_{q^{\text{old}}_{\vu}(f_i)} [\nabla_{f_i}\log p(y_i \mid f_i)],\\
   \hat{\beta}_i^\text{old} &= \myexpect_{q^{\text{old}}_{\vu}(f_i)} [ -\nabla^2_{f_i}\log p(y_i \mid f_i)].
\end{aligned} \label{eq:svgp_dual}
\end{equation}
The result follows from Eq.~(21) in \citet{adam2021dual}. A proof is given in \cref{app:dual_params} where we also derive a dual-form for the predictive distribution $q_{\vu}^\text{old}(f_i)$:
\begin{align}
   \myexpect_{q_{\vu}^\text{old}(f)}[f_i] &= \vkzs^{\T} \MKzz^{-1} \valpha_{\vu}^\text{old} , \\
   \textrm{Var}_{q_{\vu}^\text{old}(f)}[f_i]  &= \kappa_{ii} - \vkzs^\top \sqr{\MKzz^{-1} - \rnd{ \MKzz + \vBeta_{\vu}^\text{old} }^{-1} }   \vkzs. \nonumber
\end{align}
The form is strikingly similar to \cref{eq:gp_pred} but it uses two dual parameters consisting of an $m$-length vector $\valpha_\vu^\text{old}$ and $m\times m$-size matrix $\vBeta_\vu^\text{old}$, defined as
\begin{equation} 
\begin{split}
   \valpha_{\vu}^\text{old} &  = \textstyle\sum_{i \in \mathcal{D}_{\textrm{old}}}  \vkzi \, \hat{\alpha}_{i}^\text{old}, \\
   \vBeta_{\vu}^\text{old} &= \textstyle\sum_{i \in \mathcal{D}_{\textrm{old}}} \vkzi \,\hat{\beta}_{i}^\text{old} \, \vkzi^{\T} .
\label{eq:dual_sparse}
\end{split}
\end{equation}
The pair $(\valpha_{\vu}^\text{old}, \vBeta_{\vu}^\text{old})$ can be seen as `amortized' dual parameters which do not depend on the data size $n$, but rather provide a distilled compact summary through $m$ inducing inputs.
Similarly, the pair $(\hat{\alpha}_i^\text{old}, \hat{\beta}_i^\text{old})$ can be seen as an estimate of $(\alpha_i, \beta_i)$ in \cref{eq:dual_param} obtained by using $q_{\vu}^\text{old}(f_i)$ instead of the exact posterior $p(f_i|\vy^\text{old})$.

The posterior can also be expressed in a form where likelihoods are replaced by their Gaussian approximations, 
\begin{align} 
   q_{\vu}^\text{old}(\vu) &\propto e^{ \vu^\top (\MV^{\text{old}})^{-1} \vm^{\text{old}} -\half \vu^\top (\MV^{\text{old}})^{-1} \vu} \nonumber\\
      &\propto p_{\vtheta}{(\vu)} \,\,  e^{ \vu^\top (\MV^{\text{old}})^{-1} \vm^{\text{old}} -\half \vu^\top \rnd{ (\MV^{\text{old}})^{-1} - \MKzz^{-1} } \vu} \nonumber \\
      &\propto p_{\vtheta}{(\vu)} \textstyle\prod_{i\in \mathcal{D}_{\textrm{old}} } e^{-\half \hat{\beta}_i^\text{old} (\hat{y}_i^\text{old} - \va_i^\top \vu)^2}, \label{eq:sites}  
\end{align} 
where the second line is obtained by adding and subtracting $\vu^\top \MKzz^{-1}\vu/2$ and the third line is obtained by using \cref{eq:nat1,eq:nat2} and then completing the squares.
Previous works used EP to obtain such site parameters \citep{csato2002sparse, bui2017streaming}, but we obtain them with SVGP too.

\citet{adam2021dual} further show that the dual form can be used to improve hyperparameter optimization. The idea is to fix the dual parameters in \cref{eq:sites} and treat both $p_{\vtheta}(\vu)$ and $\va_i = \vkzi^\top \MKzz^{-1}$ as functions of $\vtheta$. Then, plugging \cref{eq:sites} into \cref{eq:elbo_old} gives rise to an objective which, compared to the ELBO, is better aligned with the marginal likelihood; see App.~B of \citet{adam2021dual}.
They propose a stochastic expectation-maximization procedure in which the posterior is updated by maximizing \cref{eq:elbo_old} and hyperparameters are updated using the new objective.

In the next section, we will derive a new objective for sequential learning where memory is used to mimic the batch-SVGP of \cref{eq:elbo_svgp_offline}. Similarly to \citet{bui2017streaming}, we also use the pseudo-data idea, but our approach is more straightforward and directly addresses the issues they aim to solve.

\subsection{Memory-based Objective for Sequential Learning}

Our goal is to closely mimic the batch-SVGP objective given in \cref{eq:elbo_svgp_offline}. We make two modifications to the batch objective. First, instead of $\data_{\text{old}}$, we use a subset of examples stored in  a memory set $\mathcal{M}$:
\begin{align}
   \sum_{\mathclap{i \in \mathcal{D}_{\textrm{new}} \cup \mathcal{M} }}  \mathbb{E}_{q_{\vu}(f_i)}  [\log p(y_i \mid f_i)] 
      - \dkl{}{q_\vu(\vu)}{ p_{\vtheta} {(\vu)} }.
      \label{eq:elbo_new}
\end{align}
For $\mathcal{M} = \mathcal{D}_\text{old}$, we exactly recover $\mathcal{L}_{\text{batch}}$ and, for small memory size, the error can be minimized by carefully choosing $\mathcal{M}$ as a representative set of $\data_\text{old}$.

Second, we reuse the old posterior $q_{\vu}^\text{old}(\vu)$ which contains a distilled summary of the old data in its dual parameters.
Our key idea is to express the prior in terms of the old posterior by using \cref{eq:sites},
\begin{align} 
   p_{\vtheta}{(\vu)} &\propto \frac{ q_{\vu}^\text{old}(\vu)} { \prod_{i\in \mathcal{D}_{\textrm{old}} } e^{-\half \hat{\beta}_i^\text{old} (\hat{y}_i^\text{old} - \va_i^\top \vu)^2} }  
   \propto \frac{ q_{\vu}^\text{old}(\vu) }{ \hat{p}_{\data_\text{old}}(\vu) },
   \label{eq:prior_alt}
   \end{align}
where we rewrite the denominator as a Gaussian:
\begin{align}
   \hat{p}_{\data_\text{old}}(\vu) &= \N( \vu \mid \tilde{\vy}^\text{old}, \tilde{\vSigma}^\text{old}) , \label{eq:phat}\\
   \tilde{\vy}^\text{old} &= \MKzz (\vBeta_{\vu}^\text{old})^{-1}\valpha_{\vu}^\text{old} + \vm_{\vu}^\text{old} , \\
   \tilde{\vSigma}^\text{old} &= \MKzz (\vBeta_{\vu}^\text{old})^{-1} \MKzz.  
\end{align}
A derivation is in \cref{app:MVN_correlated}.
The denominator can be seen as pseudo-data, similar to those derived in \citet{bui2017streaming}. %
\cref{eq:prior_alt} is exact whenever $q_{\vu}^{\text{old}} {(\vu)}$ is an exact stationary point of \cref{eq:elbo_old}, for example, in the very first update.
However, in sequential learning, $q_{\vu}^{\text{old}} {(\vu)}$ is almost never exact because errors can accumulate due to lack of full access to data at subsequent updates. 
To handle this, we rewrite \cref{eq:phat} where only examples in $\mathcal{M}$ are used in the denominator: \looseness-1
\begin{equation}
   \frac{ q_{\vu}^\text{old}(\vu) }{ \hat{p}_{\data_\text{old}}(\vu) } 
   \approx \frac{ \hat{q}_{\vu}^\text{old}(\vu) }{ \hat{p}_{\mathcal{M}}(\vu) },
\end{equation}
where $\hat{q}_{\vu}^\text{old}$ is an estimate of $q_{\vu}^{\text{old}} {(\vu)}$ (see \cref{methods:fast_upates}). %
The denominator is defined in similar ways but using only $\mathcal{M}$.
We will use this as our new prior which can be obtained from the dual parameters by removing contributions of $\mathcal{M}$,
\begin{align}
    \valpha_{\vu}^{\text{old} \backslash \mathcal{M}} &= \valpha_{\vu}^{\text{old}} - \textstyle \sum_{i\in\mathcal{M} } \vkzi \hat{\alpha}_i^\text{old} , \label{eq:nat1_M} \\
    \vBeta_{\vu}^{\text{old} \backslash \mathcal{M}} &= \vBeta_{\vu}^{\text{old}} - \textstyle \sum_{i\in\mathcal{M} } \vkzi \hat{\beta}_i^{\text{old}} \vkzi^\top . \label{eq:nat2_M}
\end{align}
Here, for notational simplicity, we use $(\valpha_{\vu}^{\text{old}}, \vBeta_{\vu}^{\text{old}})$ to denote the dual parameters of $\hat{q}^\text{old}_{\vu}(\vu)$.
A derivation is in \cref{app:MVN_correlated} along with expressions for the mean and covariance. 

Replacing $p_{\vtheta}(\vu)$ in \cref{eq:elbo_new} with the new prior ($\mathcal{Z}$ denotes the normalizing constant), we arrive at
\begin{align}
   \mathcal{L}^q_{\text{seq}} &= \sum_{\mathclap{i \in \mathcal{D}_{\textrm{new}} \cup \mathcal{M} }} \mathbb{E}_{q_{\vu}(f_i)}  [\log p(y_i \mid f_i)]   - \mathbb{D}_{\text{KL}} \!\bigg[\! q_\vu(\vu) \| \frac{ \hat{q}_{\vu}^{\text{old}} {(\vu)}}{\mathcal{Z} \hat{p}_{\mathcal{M}}(\vu)} \!\bigg] \!. \label{eq:elbo_new1}
\end{align}
Why do we expect the new objective in \cref{eq:elbo_new1} to be more accurate? One explanation is to see the new objective as an improved version of variational continual-learning (VCL) \citep{nguyen2017variational}, by rewriting it as
\begin{align}
   &\mathcal{L}^q_{\text{seq}} = \sum_{\mathclap{i \in \mathcal{D}_{\textrm{new}} }} \mathbb{E}_{q_{\vu}(f_i)}  [\log p(y_i \mid f_i)] - \dkl{}{q_\vu(\vu)}{ \hat{q}_{\vu}^{\text{old}} {(\vu) }} \nonumber\\
   &\,\, + \sum_{\mathclap{i \in \mathcal{M} }} \mathbb{E}_{q_{\vu}(f_i)}  [\log p(y_i \mid f_i)] - \mathbb{E}_{q_{\vu}(\vu)}[\log  \hat{p}_{\mathcal{M}}(\vu) ] .
   \label{eq:elbo_new2}
\end{align}
The first two terms are equal to the VCL objective where the estimated old posterior is used as the new prior. VCL is exact only when the old posterior is exact, which, as discussed before, is unlikely in practice. To address this issue, in the third term we add a few representative examples of $\mathcal{D}_\text{old}$ using $\mathcal{M}$. The fourth term is subtracting the pseudo-data to avoid double-counting the contributions of the examples in
$\mathcal{M}$. \looseness-1

The new objective directly addresses the issues raised by \citet{bui2017streaming} regarding hyperparameter learning, who add two KLD terms (see the last two lines of \cref{eq:elbo_bui}). This essentially uses the ratio $q_{\vu_{\textrm{old}}}(\vu_{\textrm{old}}) /p_{\vtheta_{\textrm{old}}}(\vu_{\textrm{old}}) $, which is proportional to the pseudo-data. 
In contrast, in our objective, we use the exact likelihood (the third term in \cref{eq:elbo_new2}) in addition to the pseudo-data (the fourth term). The two terms are added on top of the VCL objective where $\hat{q}_{\vu}^\text{old}(\vu)$ is used. %

In the limit of $\mathcal{M} = \mathcal{D}_\text{old}$, we have $\mathcal{L}_\text{seq}^q = \mathcal{L}_{\text{batch}}$ (assuming $\vZ$ and $\vtheta$ to be the same): In this case, the first terms of both \cref{eq:elbo_new1} and \cref{eq:elbo_svgp_offline} are equal; the second terms can also be seen to be equal using \cref{eq:prior_alt}.
By using a good representative memory of the past data, we expect to mimic the batch-SVGP objective.
Using memory to improve sequential learning is unique to our approach.

\begin{figure*}[t!]
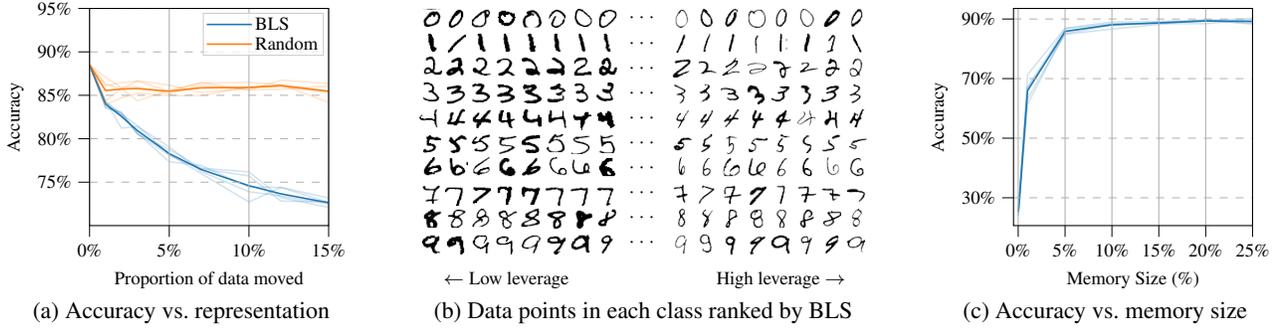

	\scriptsize
	\centering
	\setlength{\figureheight}{.35\columnwidth}	
	\setlength{\figurewidth}{1.1\figureheight}
	\begin{subfigure}[b]{.28\textwidth}
		\pgfplotsset{axis on top, scale only axis,ytick={.75,.8,.85,.9,.95, 1.},yticklabels={75\%,80\%,85\%, 90\%,95\%,100\%},xmin=0,xmax=15,ymin=.7,ymax=.95,xtick={0,5, 10,15}, xticklabels={0\%,5\%, 10\%,15\%}, y grid style={line width=.1pt, draw=gray!10,dashed},x grid style={solid},grid,ylabel style={yshift=-1em}}
		
\begin{tikzpicture}

\definecolor{darkgray176}{RGB}{176,176,176}
\definecolor{darkorange25512714}{RGB}{255,127,14}
\definecolor{lightgray204}{RGB}{204,204,204}
\definecolor{steelblue31119180}{RGB}{31,119,180}

\begin{axis}[
height=\figureheight,
legend cell align={left},
legend style={fill opacity=0.8, draw opacity=1, text opacity=1, draw=lightgray204},
tick align=outside,
tick pos=left,
width=\figurewidth,
x grid style={darkgray176},
xlabel={Proportion of data moved},
xmin=-0, xmax=15,
xtick style={color=black},
y grid style={darkgray176},
ylabel={Accuracy},
ytick style={color=black}
]
\addplot [semithick, steelblue31119180]
table {%
0 0.884945007856021
0 0.884945007856021
0 0.884945007856021
1 0.839859899732161
2 0.825752265061835
3 0.80937440214272
5 0.782941491577882
7 0.764600189721556
10 0.745931330034182
12 0.736576420056947
15 0.726271148609437
};
\addlegendentry{BLS}
\addplot [semithick, darkorange25512714]
table {%
0 0.884945007856021
0 0.884945007856021
0 0.884945007856021
1 0.855504429640821
2 0.8571390781033
3 0.85775141891461
5 0.85468722099875
7 0.858612800624965
10 0.859058211315749
12 0.861358042565513
15 0.854720771394134
};
\addlegendentry{Random}
\addplot [semithick, steelblue31119180, opacity=0.2, forget plot]
table {%
0 0.884945007856021
0 0.884945007856021
0 0.884945007856021
1 0.835862921502644
2 0.83102969380332
3 0.804986926854155
5 0.787691208856616
7 0.766140282350315
10 0.76189990306617
12 0.732445345301868
15 0.720905316726932
};
\addplot [semithick, darkorange25512714, opacity=0.2, forget plot]
table {%
0 0.884945007856021
0 0.884945007856021
0 0.884945007856021
1 0.870750635258567
2 0.85873950135573
3 0.859894139404375
5 0.851496934706268
7 0.861503264326767
10 0.856231824906893
12 0.867429178128469
15 0.863622159725012
};
\addplot [semithick, steelblue31119180, opacity=0.2, forget plot]
table {%
0 0.884945007856021
0 0.884945007856021
0 0.884945007856021
1 0.835862921502644
2 0.829574763573838
3 0.806389898603405
5 0.773584905660377
7 0.769209307516322
10 0.744349778072547
12 0.743931277448
15 0.726663988214812
};
\addplot [semithick, darkorange25512714, opacity=0.2, forget plot]
table {%
0 0.884945007856021
0 0.884945007856021
0 0.884945007856021
1 0.838678662179795
2 0.847166192712122
3 0.852369109112939
5 0.849592286173442
7 0.849729367780816
10 0.859445946635376
12 0.859804063510448
15 0.854292219097362
};
\addplot [semithick, steelblue31119180, opacity=0.2, forget plot]
table {%
0 0.884945007856021
0 0.884945007856021
0 0.884945007856021
1 0.843279994505872
2 0.81224786720455
3 0.814106243224284
5 0.790607701922505
7 0.762401651693544
10 0.738992908525075
12 0.734182713189518
15 0.724744431052185
};
\addplot [semithick, darkorange25512714, opacity=0.2, forget plot]
table {%
0 0.884945007856021
0 0.884945007856021
0 0.884945007856021
1 0.863264885653458
2 0.857416837510747
3 0.866653912378037
5 0.861615380036903
7 0.853300597064896
10 0.858527626141523
12 0.860431446358766
15 0.852729788848712
};
\addplot [semithick, steelblue31119180, opacity=0.2, forget plot]
table {%
0 0.884945007856021
0 0.884945007856021
0 0.884945007856021
1 0.842387198681409
2 0.827590767806362
3 0.806198584273962
5 0.780846378191774
7 0.765247475029295
10 0.757257282791694
12 0.728198446020945
15 0.727378242042766
};
\addplot [semithick, darkorange25512714, opacity=0.2, forget plot]
table {%
0 0.884945007856021
0 0.884945007856021
0 0.884945007856021
1 0.860929881189479
2 0.864625355465908
3 0.866207512276003
5 0.85465150883876
7 0.863902684002009
10 0.86495586959849
12 0.862168814246417
15 0.8420159814294
};
\addplot [semithick, steelblue31119180, opacity=0.2, forget plot]
table {%
0 0.884945007856021
0 0.884945007856021
0 0.884945007856021
1 0.841906462468237
2 0.828318232921103
3 0.815190357757796
5 0.781977263258139
7 0.760002232018303
10 0.727156777715423
12 0.744124318324405
15 0.731663765010491
};
\addplot [semithick, darkorange25512714, opacity=0.2, forget plot]
table {%
0 0.884945007856021
0 0.884945007856021
0 0.884945007856021
1 0.843898083922808
2 0.857747503471993
3 0.843632421401696
5 0.856079995238379
7 0.864628089950338
10 0.856129789296464
12 0.856956710583466
15 0.860943707870184
};
\end{axis}

\end{tikzpicture}
		\caption{Accuracy vs.\ representation}
		\label{fig:mnist-bls-random}
	\end{subfigure}
	\hfill
	\begin{subfigure}[b]{.4\textwidth}	
		\centering	
		\begin{tikzpicture}
			\newlength{\digitwidth}     
			\setlength{\digitwidth}{.115\figureheight}
			\def\datapath{./fig/split_mnist/digit_bls}

            \foreach \i in {0,...,9} {

              \foreach \x [count=\j] in {39,38,37,35,31,27,23,20}
				\node[minimum width=\digitwidth,minimum height=\digitwidth, inner sep=0]at ($({\digitwidth*\j},{-\digitwidth*\i})$)
				{\includegraphics[width=\digitwidth]{\datapath/\i/\x.png}};

              \node at ($({\digitwidth*9.5},{-\digitwidth*\i})$) {$\cdots$};

              \foreach \x [count=\j from 11] in {19,15,11,7,3,2,1,0}
				\node[minimum width=\digitwidth,minimum height=\digitwidth, inner sep=0]at ($({\digitwidth*\j},{-\digitwidth*\i})$)
				{\includegraphics[width=\digitwidth]{\datapath/\i/\x.png}};
				
			}

			\node at ($({\digitwidth*4},{-\digitwidth*10.5})$) {\scriptsize $\leftarrow$ Low leverage};
			\node at ($({\digitwidth*15},{-\digitwidth*10.5})$) {\scriptsize High leverage $\rightarrow$};
			
		\end{tikzpicture}%
		\caption{Data points in each class ranked by BLS}
		\label{fig:mnist-bls}
	\end{subfigure}
	\hfill
	\begin{subfigure}[b]{.28\textwidth}
		\pgfplotsset{axis on top, scale only axis,ytick={.2, .3, .5, .7 , .9, .95},yticklabels={,30\%, 50\%,70\%, 90\%,100\%},xmin=0,xmax=15,ymin=.7,ymax=.95,xtick={0, 5, 10, 15, 20, 25}, xticklabels={0\%, 5\%, 10\%, 15\%, 20\%, 25\%}, y grid style={line width=.1pt, draw=gray!10,dashed},x grid style={solid},grid,ylabel style={yshift=-1em}}
		
\begin{tikzpicture}

\definecolor{darkgray176}{RGB}{176,176,176}
\definecolor{steelblue31119180}{RGB}{31,119,180}

\begin{axis}[
height=\figureheight,
tick align=outside,
tick pos=left,
width=\figurewidth,
x grid style={darkgray176},
xlabel={Memory Size (\%)},
xmin=-0.5, xmax=25,
xtick style={color=black},
y grid style={darkgray176},
ylabel={Accuracy},
ymin=0.206624053706613, ymax=0.935427081845451,
ytick style={color=black}
]
\addplot [semithick, steelblue31119180]
table {%
0 0.253578060277103
0.05 0.264462219682902
1 0.659562919582917
5 0.858563062419654
10 0.88061705470647
20 0.893215254963577
25 0.892155377803171
};
\addplot [semithick, steelblue31119180, opacity=0.2]
table {%
0 0.258748750178546
0.05 0.261605484930724
1 0.631338380231396
5 0.85080702756749
10 0.879302956720469
20 0.895443508070276
25 0.890944150835595
};
\addplot [semithick, steelblue31119180, opacity=0.2]
table {%
0 0.240608484502214
0.05 0.260319954292244
1 0.613840879874304
5 0.856949007284674
10 0.882802456791887
20 0.893729467218969
25 0.886230538494501
};
\addplot [semithick, steelblue31119180, opacity=0.2]
table {%
0 0.279817168975861
0.05 0.279460077131838
1 0.669618625910584
5 0.867161834023711
10 0.884873589487216
20 0.884087987430367
25 0.899
};
\addplot [semithick, steelblue31119180, opacity=0.2]
table {%
0 0.23975146407656
0.05 0.260248535923439
1 0.669904299385802
5 0.848735894872161
10 0.865947721754035
20 0.897943150978432
25 0.882302528210256
};
\addplot [semithick, steelblue31119180, opacity=0.2]
table {%
0 0.248964433652335
0.05 0.260677046136266
1 0.713112412512498
5 0.869161548350236
10 0.890158548778746
20 0.89487216111984
25 0.902299671475503
};
\end{axis}

\end{tikzpicture}
		\caption{Accuracy vs.\ memory size}
		\label{fig:split-mnist-mem-acc}
	\end{subfigure}
	\hfill
  \caption{Ablation study on {\em split MNIST} to explore the benefits of memory and Bayesian leverage score (BLS). (a)~Evolution of test set accuracy of our method as we move data from training set to test set based on BLS ranking vs.\ randomly. (b)~Digits from the training set with lowest$\leftrightarrow$highest BLS, high BLS digits are more unusual and more difficult to learn than digits with low BLS score. (c)~Evolution of test set accuracy as the memory size is increased; memory size of just $5\%$ achieves satisfactory performance.}
\end{figure*}

\subsection{Inference using the Memory-Based Objective}

\label{methods:fast_upates}
We will optimize the new objective in \cref{eq:elbo_new1} by using the Bayesian learning rule (BLR) of \citet{khan2021BLR} which is a natural-gradient descent algorithm. This method is also used by \citet{adam2021dual} who write the update in a natural-parameter form. We will instead use a form where we update the estimate of the dual pair $(\valpha_{\vu}^{(t)}, \vBeta_{\vu}^{(t)})$ at each iteration $t$. A detailed derivation is given in \cref{app:ngd_deriv} and
below we give the final update,
\begin{equation}\label{eq:updates_iter}
    \begin{aligned}
       \valpha_\vu^{(t)} &\leftarrow (1 -\rho)\valpha_\vu^{(t-1)}  + \rho \Big( \valpha_{\vu}^{\text{old} \backslash \mathcal{M}} + \sum_{\mathclap{i\in\data_\text{new}}} \vkzi \hat{\alpha}_i^{(t)} \Big) , \\
       \vBeta_{\vu}^{(t)} &\leftarrow (1 - \rho)\vBeta_{\vu}^{(t-1)} + \rho \Big( \vBeta_{\vu}^{\text{old} \backslash \mathcal{M}} + \sum_{\mathclap{i\in\data_\text{new}}} \vkzi \hat{\beta}_i^{(t)} \vkzi^\top \Big).
\end{aligned}
\end{equation}
The update adds the contribution of the new data using the dual parameters $(\hat{\alpha}_i^{(t)}, \hat{\beta}_i^{(t)})$ which are defined similarly to \cref{eq:nat1,eq:nat2} but by using the most recent posterior $q_{\vu}^{(t-1)}(f_i)$. The mean and covariance of the posterior needed to compute the expectations are obtained as follows,
\begin{equation}
   \vm^{(t)} \leftarrow \valpha_{\vu}^{(t)}, ~~
   \MV^{(t)} \leftarrow \big( \MKzz^{-1} \vBeta_{\vu}^{(t)} \MKzz^{-1} + \MKzz^{-1} \big)^{-1}.
   \label{eq:meancov_itert_1}
\end{equation}

\subsection{Learning of Hyperparameter and Inducing Inputs}
For hyperparameter learning, we will use the new ELBO derived by \citet{adam2021dual} who show faster convergence; also see the recent work by \citet{li2023improving}. As before, we will derive it by assuming that $\mathcal{D}_\text{old}$ is available and then we will make approximations to handle the sequential case.

We follow the procedure given in \citet[App.~B]{adam2021dual}.
Let $q_{\vu}^\text{new} (\vu)$ denote the solution obtained by maximizing \cref{eq:elbo_svgp_offline}. The idea is to take its dual form and fix $(\hat{y}_i^\text{new}, \hat{\beta}_i^\text{new})$ while treating both $p_{\vtheta}(\vu)$ and $\va_i$ as functions of $\vtheta$:
\begin{align} 
   q_{\vu}^\text{new}(\vu; \vtheta) &\propto p_{\vtheta}{(\vu)} \textstyle\prod_{i\in \mathcal{D} } e^{-\half \hat{\beta}_i^\text{new} (\hat{y}_i^\text{new} - \vu^\top \va_i^{\vtheta} )^2}, \label{eq:sites_new}  
\end{align} 
where $\data = \data^\text{new} \cup \data^\text{old}$ and we have indicated explicit dependency on $\vtheta$ by denoting $\va_i^{\vtheta}$ and $q_{\vu}^\text{new}(\vu; \vtheta)$.
 Then, plugging \cref{eq:sites_new} in \cref{eq:elbo_old} gives the following objective \citep[see App.~B in][for details]{adam2021dual}:
\begin{align}
   \mathcal{L}_{\textrm{seq}}^{\vtheta} &=  \log \mathcal{Z}(\vtheta) + c(\vparam) ,
   \label{eq:elbo_svgp_hyps}\\%
   c(\vtheta) &= \sum_{\mathclap{i \in \data}}  \mathbb{E}_{q(f_i;\vtheta)}[ \log p(y_i \mid f_i)] - \mathbb{E}_{q(\vu;\vtheta)}[\log \hat{p}_{\data } (\vu)], \nonumber
\end{align}
\vspace{-5ex}
\begin{multline*}
   \log \mathcal{Z}(\vtheta) = -\frac{1}{2} \log \left| \MKzz^{\vtheta}  (\vBeta_{\vu}^{\text{new}})^{-1} \MKzz^{\vtheta} + \MKzz^{\vtheta} \right| \nonumber\\
   - \frac{1}{2} \vb^\top \big( \vBeta_{\vu}^{\text{new}} + \MKzz^{\vtheta} \big)^{-1} \vb + \text{const}. 
\end{multline*}
Here, we keep $\vB_{\vu}^\text{new}$ fixed, and treat everything else as a function of $\vtheta$. We define $\vb = (\vB_{\vu}^\text{new})^{-1} \valpha_{\vu}^\text{new} + \vm^\text{new}$ which contains the part in $\tilde{\vy}^\text{new}$ that are fixed.

For the sequential setting, we modify the first term in $c(\vtheta)$ by using a stochastic approximation for the sum over $\mathcal{D}_{\textrm{old}}$. We approximate it by using $\mathcal{M}$ and scale it to get an unbiased estimate of the gradient,
\begin{align*}
   &\sum_{\mathclap{i \in \data^\text{new}}}  \mathbb{E}_{q(f_i;\vtheta)}[ \log p(y_i \mid f_i)] \\
   & + \frac{n_\textrm{old}}{n_{\mathcal{M}}} \sum_{\mathclap{i \in \mathcal{M} }}  \mathbb{E}_{q(f_i;\vtheta)}[ \log p(y_i \mid f_i)].
\end{align*}
Here, $n_\textrm{old}$ and $n_{\mathcal{M}}$ are the size of $\data_\text{old}$ and $\mathcal{M}$ respectively. This is similar to the approach of \citet{titsias2019functional}.
To update the inducing inputs, we use the pivoted-Cholesky method of \citet{burt2020convergence}. A similar method is used by \citet{maddox2021conditioning} but in combination with gradient based optimization. We find that our framework gives much better performance using the pivoted-Cholesky alone, and we do not need any gradient based optimization of inducing points. We concatenate $[\MZ_{\textrm{old}},\MX_{\textrm{new}}]$ and use pivoted-Cholesky to get the new inducing points
$\MZ_{\textrm{new}}$. Having moved the inducing points to $\MZ_{\textrm{new}}$, we now need to adjust the variational parameters which are still defined over the old inducing points $\MZ_{\textrm{old}}$.
We adjust this using a projection matrix $\MP = \MK_{\mbf{Z}_{\textrm{new}},\mbf{Z}_{\textrm{old}}}\MK_{\mbf{Z}_{\textrm{old}},\mbf{Z}_{\textrm{old}}}^{-1}$ to get the new dual parameters,
\begin{equation} \label{eq:update_lambda}
\valpha_{\vu} \leftarrow \MP \valpha_{\vu}  \quad \text{and} \quad \vBeta_{\vu}  \leftarrow \MP \vBeta_{\vu} \MP^{\T}. 
\end{equation}

\subsection{Memory Selection using Bayesian Leverage Score}
\label{methods:bls}

Another key contribution of our work is to select and update a memory of the past data to improve learning. We present a new score called the Bayesian leverage score (BLS) to actively build and update the memory. The score extends the classical ridge leverage score \citep[RLS,][]{alaoui2015fast}, which is commonly used to select subsets for linear regression. We generalise it to the SVGP case, and use it to build a memory for sequential learning. \cref{fig:mnist-bls} gives an example of ranking data instances by BLS (details in \cref{sec:mnist}).%

\begin{algorithm}[t!]
	\caption{Dual-SVGP with memory} 
	\label{alg:BO/AL}
	\begin{algorithmic}[1] 
      \State Initialize $\valpha_{\vu}, \vBeta_{\vu}, \MZ_{\textrm{old}}, \vtheta,  \mathcal{M}$ 
	  \For{each new data}
      		\State Observe new data $(\vy_{\textrm{new}}, \MX_{\textrm{new}})$
	  		\State Update $\MZ$ using pivoted-Cholesky on $[\MZ_{\textrm{old}}, \MX_{\textrm{new}}]$ 
      		\State Project old $(\valpha_{\vu}, \vBeta_{\vu})$ to new $\MZ$ using \cref{eq:update_lambda}
      \State Update $(\valpha_{\vu}, \vBeta_{\vu})$ by using \cref{eq:updates_iter,eq:meancov_itert_1}
      \State Update $\vtheta$ by optimizing \cref{eq:elbo_svgp_hyps}
		\State Update memory $\mathcal{M}$ using the method from \cref{methods:bls} 
		\EndFor
	\end{algorithmic}
\end{algorithm}

Consider the following linear model,
$y_i = \va_i^\top \vu + \epsilon_i$,
where $y_i$ are the observations, $\va_i$ are the features, $\vu$ is the parameter vector, and $\epsilon_i \sim \N(0,\sigma^2)$. If we consider Bayesian linear regression in this model, \ie we place a prior over the weights $\N(\vu; \vzeros, \MKzz)$, then
\begin{equation} \label{eq:ridge}
 p(\vu \mid \vy) \propto \gauss(\bm{0}, \MKzz) \textstyle\prod_{i=1}^n e^{-\frac{1}{2\sigma^2}(y_i - \va_i^\top \vu)^2 },
\end{equation}
highlighting the similarity to \cref{eq:sites}. The ridge leverage score for the $i$\textsuperscript{th} example is defined as 
\begin{equation}
   h_i^{\text{RLS}} = \va_i^\top \rnd{\frac{1}{\sigma^2} \MA \MA^\top +  \MKzz)^{-1}}^{-1} \va_i, 
\end{equation}
where $\MA = (\va_1, \ldots, \va_n)$. Examples with high leverage scores are often far away from other observations and have high predictive uncertainty.

Comparing the above linear model to  \cref{eq:sites}, we see that the SVGP posterior is equivalent to the posterior of a linear model with a major difference: the noise is now heteroscedastic with variance $1/\hat{\beta}_i^\text{new}$ for the $i$\textsuperscript{th} example. This motivates the following Bayesian generalisation of RLS, which we refer to as the Bayesian leverage score (BLS),
\begin{equation}
   h_i^{\text{BLS}} = \va_i^\top \rnd{\MA \, \diag(\hat{\vbeta}^\text{new}) \, \MA^\top + \MKzz^{-1}}^{-1} \va_i .
\label{eq:gvim}
\end{equation}
We do not have to invert $\MKzz$ explicitly. The score can be computed without any additional cost by rewriting it in terms of predictive variance $\hat{v}_i$ of $q_{\vu}(f_i)$, that is,
\begin{equation} \label{eq:bls_score}
   h^{\textrm{BLS}}_i = \hat{\beta}_i^{\text{new}} \hat{v}_i^{\text{new}}.
\end{equation}
For Gaussian likelihoods, BLS in \cref{eq:bls_score} reduces to RLS because $\hat{\beta}_i = 1/\sigma^2$, but unlike RLS, the BLS also applies to non-Gaussian likelihoods.
When selecting the memory, we want representative examples. We achieve this by sampling memory points from the new batch weighted by the BLS score. This biases our memory towards more difficult examples whilst covering the typical set, which we find improves performance. The final algorithm is given in \cref{alg:BO/AL}.

\section{Experiments}
\label{sec:experiments}
We perform a range of experiments to show the capability of the proposed method on various sequential learning problems. %
For sequential decision-making tasks, we apply our method to Bayesian Optimization (BO) and Active Learning (AL) (\cref{sec:BO_AL}). Building on \citet{chang2022fantasizing}, we show how \emph{`fantasy'} batch acquisition functions can be built for simple acquisition functions and demonstrate the effectiveness on the `lunar landing' BO problem and the `hotspot modelling' AL problem.\looseness-1

In streaming tasks, data comes in small batches, where the total number of data points is unknown, and all SVGP parameters should be optimized. This limits the use of the methods by \citet{maddox2021conditioning} and \citet{kapoor2021variational}.
Therefore, we compare the proposed method against \citet{bui2017streaming} on streaming UCI tasks and on a real-world robot task (\cref{sec:streaming}). 
Finally, we consider the \emph{split-MNIST} continual learning problem, where the data is non-stationary, but the total number of tasks and data points are known beforehand. Thus, in \cref{sec:mnist} we compare the proposed method to \citet{kapoor2021variational} and \citet{bui2017streaming}. Additionally, we include a study on the benefits of our BLS score over a random selection of memory.

\begin{figure}[b!]
	\centering\scriptsize
	\def\datapath{./fig/lunar/}
	\centering
	\setlength{\figurewidth}{.4\columnwidth}
	\setlength{\figureheight}{1.1\figurewidth}
	\pgfplotsset{scale only axis,axis on top, yticklabel style={rotate=90,anchor=base,yshift=1pt},ylabel style={yshift=-3.5em}}
	\begin{subfigure}[b]{.48\columnwidth}
		\centering
		\tiny
		\input{fig/lunar_lander_reward_obs}
	\end{subfigure}
	\hfill
	\begin{subfigure}[b]{.48\columnwidth}
		\centering\normalsize
		\resizebox{\textwidth}{!}{%
			\begin{tikzpicture}[inner sep=0]
				\node[anchor=north west] at (0,0) {\includegraphics[width=6cm,trim=120 100 0 0,clip]{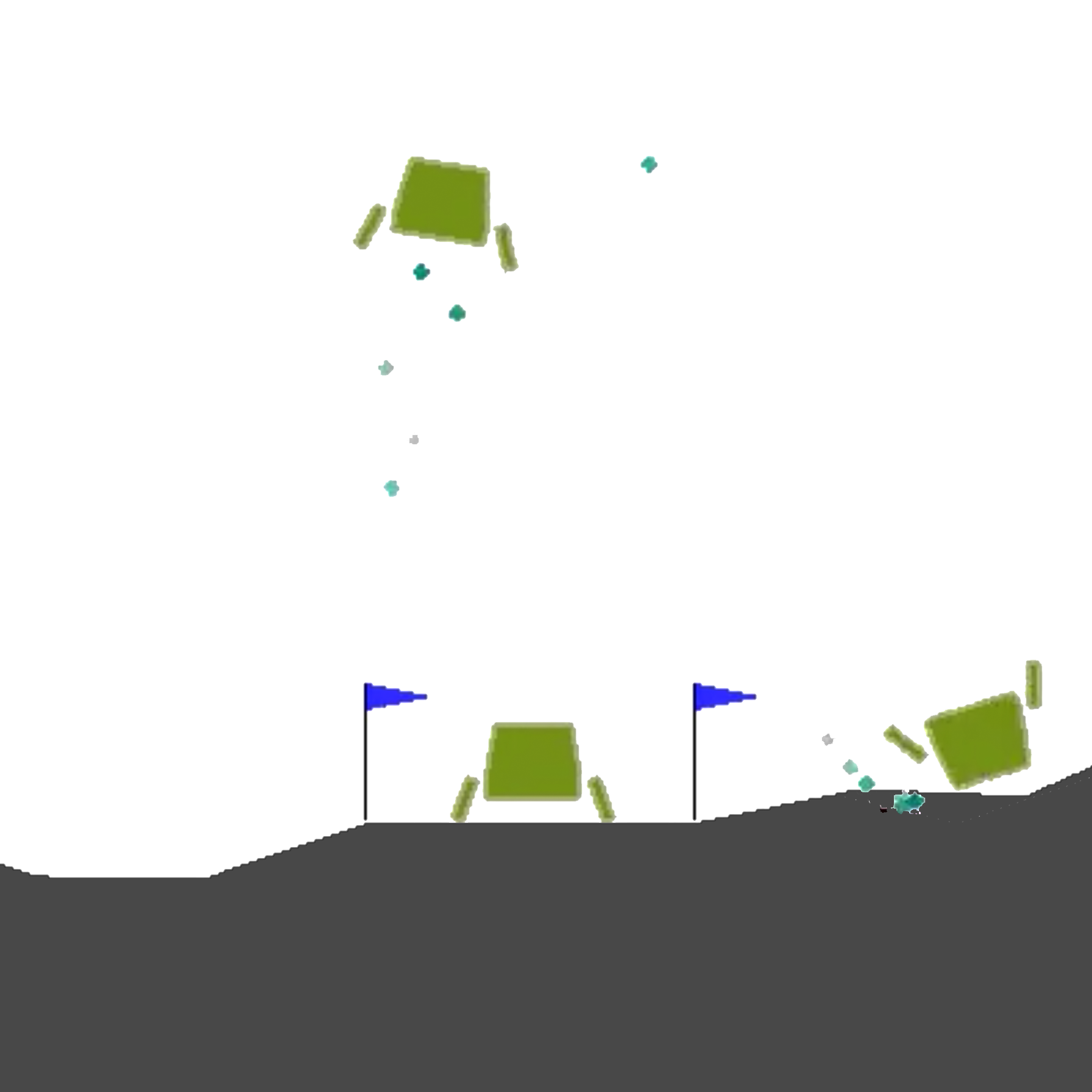}};

				\coordinate (a) at (1.9,-1.3);

				\draw[black,-latex',->,thick] (a) -- ++(0,-1);
				\draw[black,-latex',->,thick] (a) -- ++(1,0);

				\node[align=left,font=\small] at ($(a) + (2.7,-1.25)$) {
					State variables: \\
					$\bullet$~position: $x,y$ \\
					$\bullet$~velocity: $v_x, v_y$ \\
					$\bullet$~orientation: $\omega, \dot{\omega}$ \\
					$\bullet$~indicator variables\\
					$\bullet$~thruster states};
				
				\node at ($(a) + (0,.5)$) {Random initial state};
				\node at ($(a) + (.5,-4.4)$) {\color{white}Success};
				\node at ($(a) + (2.7,-4.3)$) {\color{white}Failure};
				\node[rotate=90] at ($(a) + (-1.8,-1.8)$) {Stochasticity from wind};
				\draw[->,blue!50,decoration={snake}, decorate] ($(a) + (-1.7,-1.4)$) -> ++(1,0);
				\draw[->,blue!50,decoration={snake}, decorate] ($(a) + (-1.7,-2.4)$) -> ++(1,0);
				
		\end{tikzpicture}}\\[.5cm]~
		\captionsetup{justification=centering}
	\end{subfigure}
	\vspace*{-2em}
	\caption{Bayesian optimization on a lunar landing setup. The goal is to successfully land on the surface between the flags. Compared against the non-batch solution and a batch solution using parallel Kriging  \citep{Ginsbourger2010},  the proposed approach achieves higher reward in less function evaluations.\looseness-1}
	\label{fig:lunar_landing}
\end{figure}

\subsection{Bayesian Optimization and Active Learning}
\label{sec:BO_AL}
The {\em lunar lander problem} is a challenging optimization problem that aims to successfully land a vehicle inside a specific region of a lunar surface \cite{henry:bosh}. Here, every action performed by the lander results in a reward. The aim is to optimize the total reward. Various environmental components add to the model's stochasticity, making it a challenging problem. The setup with the sources of stochasticity and the multiple states of the lander is shown in \cref{fig:lunar_landing}. The search space spanning over $\mathbb{R}^{12}$ is high-dimensional, making the task challenging. However, building a batch based on fantasy points can help overcome the difficulty in BO. These fantasy points are simulated points $(\vx_i, \vy_i)$ obtained from the acquisition function $\gamma(\cdot)$ by $\vx_i = \arg \max_{\vx} \gamma(\vx)$ and sampling the corresponding $\vy_i$ from the current model. We then condition the fantasized point into the posterior using our dual updates, and repeat the procedure for subsequent query points, thereby constructing a batch of query points (see \cref{alg:BO_fast_conditioning} in \cref{app:BO}).

\begin{figure}[b!]
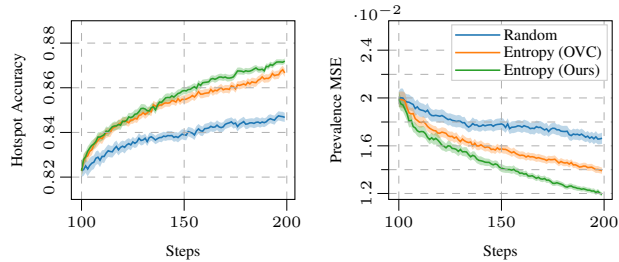

    \vspace*{-1.5em}
	\centering\tiny
	\setlength{\figurewidth}{.36\columnwidth}
	\setlength{\figureheight}{.8\figurewidth}
	\pgfplotsset{axis on top, scale only axis,legend style={font=\tiny},grid style={line width=.1pt, draw=gray!10,dashed},grid,y tick label style={rotate=90,font=\tiny},ylabel style={yshift=-2.5em},x tick label style={font=\tiny},
	y tick scale label style={rotate=-90,xshift=-2em}}
	\begin{subfigure}[b]{.48\columnwidth}
		\centering
		\pgfplotsset{legend pos=north west}
		\input{fig/hotspots/hotspots-results-acc.tex}
	\end{subfigure}
	\hfill
	\begin{subfigure}[b]{.48\columnwidth}
		\centering
        \vspace{-5ex}
		\pgfplotsset{ytick={0.012, 0.016, ..., 0.026},minor ytick={0.014, 0.018, ..., 0.024},grid=minor}
		\input{fig/hotspots/hotspots-results-mse.tex}
	\end{subfigure}
	\vspace*{-2.5em}
	\caption{Schistosomiasis hotspot modelling experiment of \citet{maddox2021conditioning}'s online variational conditioning method (OVC). We report mean $\pm$ standard error over 50 seeds, showing better performance of the proposed method.}
	\label{fig:hotspots}
\end{figure}

\begin{figure*}[t!]
	\centering\footnotesize
	\setlength{\figurewidth}{.20\textwidth}
	\setlength{\figureheight}{0.75\figurewidth}
	\begin{subfigure}{\textwidth}
		\centering
		\begin{tikzpicture}[inner sep=0]
			\node[minimum width=\figurewidth] (robot) at (0\figurewidth,0) {
				\includegraphics[width=.66\figurewidth]{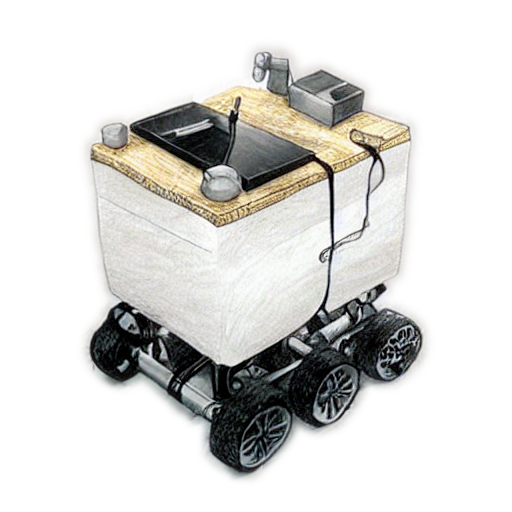}};
			\draw[black] ($(robot.center) + (5mm,5mm)$) -- ++(5mm,5mm) -- node[anchor=south]{\tiny Magnetometer} ++(1.4cm,0); 

			\foreach \t [count=\i] in {1, 2, 3, 4} {
				\node (robot-full-\i) at (\i\figurewidth,0) {
					\includegraphics[width=\figurewidth]{./fig/robot-full-\t}};
				\node[below of=robot-full-\i] {\scriptsize After path \t};
			}
		\end{tikzpicture}\\[1em]
		\caption{Experiment~\#1: Sequential estimation of magnetic field anomalies in a fixed-size domain with multiple observation paths}
		\label{fig:robot-a}
	\end{subfigure}\\[1em]
	\begin{subfigure}{\textwidth}
		\centering
		\begin{tikzpicture}[inner sep=0]
			\foreach \t [count=\i] in {0,5,10,15,20} {
				\node (robot-\i) at (\i\figurewidth,0) {
					\includegraphics[width=\figurewidth]{./fig/robot\t}};
				\node[below of=robot-\i] {\scriptsize Step \#\t};
			}

			\node at (robot-1.north) {\tiny RMSE: {\color{black}Ours} vs.\ {\color{Blue}\citeauthor{bui2017streaming}}};
			\foreach \rmseBui/\rmseOurs [count=\i from 2] in {8.31/7.79, 10.9/7.75, 9.90/7.40, 9.79/7.38} {
				\node at (robot-\i.north) {\tiny {\color{black}\rmseOurs}\ vs.\ {\color{black}\rmseBui}};
			}

			\node[anchor=south] at ($(robot-5.center) + (12mm,-4mm)$) {\includegraphics[width=3.0mm]{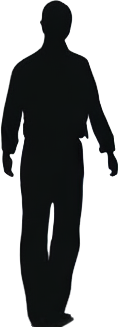}};

			\node[anchor=south] at ($(robot-1.center) + (-6mm,2mm)$) {\includegraphics[width=.07\figurewidth]{./fig/diddyborg}};
			
		\end{tikzpicture}\\[1em]
		\caption{Experiment~\#2: Maintaining representation in a sequential setting in an ever-expanding domain}
		\label{fig:robot-b}
	\end{subfigure}\\[-6pt]
	\caption{Sequential estimation of anomalies by the proposed method in the local ambient magnetic field strength (\SI{10}{\micro\tesla}~\protect\includegraphics[width=1cm,height=.65em]{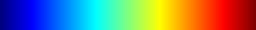}~\SI{90}{\micro\tesla}) as mapped by a small wheeled robot. Two experiments for learning the hyperparameters and representation: (a)~Complete trajectories are received and the model does not have access to the previous trajectories. (b)~Data is received during the exploration of the space requiring the method to spread out a fixed number of inducing points. Each step shows RMSE values comparing our method to {\color{black}\citet{bui2017streaming}}.}
	\label{fig:robot}
\end{figure*}

We model the problem using a regression model that aims to maximize the reward and a classification model that models whether the landing was successful. The acquisition function is a product of the Expected Improvement of the regression model and the predictive mean of the classification model. In both the models, we make a batch of  fantasy points. For details see \cref{appendix:lunar}. %

The presence of the non-Gaussian likelihood in the classification model prevents the use of many advanced batch acquisition functions which exploit properties unique to Gaussian likelihoods, such as q-KG \citep{wu2016parallel}. Our method is agnostic to the likelihood and also allows batching of fantasy points in this challenging setup. We run two baselines: a non-batch version of the proposed method and a standard batch BO method proposed by \citet{Ginsbourger2010}. Each method gets the same set of 24 initial data points and optimizes over 90 function evaluations. Both batch methods build batches of three query points. We run the experiment with five random initial observations and visualize individual and average reward curves over the BO iterations in \cref{fig:lunar_landing}. Our method improves over the non-batch setting and over the batch method of \citet{Ginsbourger2010}, showing the usefulness of our batch method and robustness to the non-Gaussian likelihood.\looseness-1

We also consider the \emph{schistosomiasis hotspot modelling} active learning problem from \citet{maddox2021conditioning} \citep[originally based on][]{andrade2020finding}. In the same SVGP model and setup (we use \citeauthor{maddox2021conditioning}'s code, though we turn off their tempering, which improves all methods), our method for updating the posterior leads to clear improvement of MSE and slightly improved accuracy (\cref{fig:hotspots}) whilst being faster.
On the same GPU, \citeauthor{maddox2021conditioning}'s online variational conditioning (OVC) method took on average \SI{85}{\second} per step (standard deviation \SI{11}{\second}); ours took on average \SI{62}{\second} (standard deviation \SI{7}{\second}).

\begin{table}[t!]
    \vspace*{-.5em}
	\centering\footnotesize
	\caption{UCI data sets: negative log predictive density, mean (standard deviation) over 10-fold cross-validation, lower is better.}
	\label{tbl:uci_experiment_full}
	\setlength{\tabcolsep}{7pt}
	\begin{tabularx}{\columnwidth}{l c c c} 
		\toprule
		\textbf{Data set} & \textbf{Offline} & \textbf{Ours} & \textbf{\citeauthor{bui2017streaming}} \\
		\midrule
		\sc Elevators	  		   & .50 (.01) & .57 (.02) & .57 (.02)  \\
		\sc Bike 					  & .24 (.03) & .44 (.03)  & .49 (.03) \\
		\sc Mammographic   & .40 (.04) & .41 (.05)   & .42 (.05)\\
		\sc Bank 			         & .25 (.03) & .26 (.03) & .28 (.04)\\
		\sc Mushroom	 	   & .00 (.00)  & .02 (.00) & .08 (.01)\\
		\sc Adult					 &  .32 (.00)	& .35 (.01) & .36 (.00) \\
		\bottomrule
	\end{tabularx}
\end{table}

\subsection{Streaming Tasks}
\label{sec:streaming}
Under streaming scenarios, where data comes in small batches and the total number of data points is unknown, we include two different experiment setups: UCI tasks and the so-called banana data set, as well as a real-world robot data set for mapping magnetic anomalies in an indoor space. 

\paragraph{UCI and Banana} We consider a setup where the {\em banana} data set and UCI \citep{Dua:2019} data sets are converted into streaming setups by splitting them into sets, ${\mathcal{D} {=} \{\mathcal{D}_1, \mathcal{D}_2, \ldots \}}$. The model at each step has access only to the current set $\mathcal{D}_{k}$. For UCI data sets, we sort with respect to the the first input dimension. \cref{fig:teaser} (top) shows the posterior obtained by the proposed method on the banana data set. Details about the setup and comparison with other methods can be found in \cref{appendix:banana}. %
In UCI data sets, we experiment with both regression and classification setups. The offline model has access to the whole data and can take multiple passes. Therefore, the offline model is used as a gold-standard baseline and we compare our method with \citet{bui2017streaming}. \cref{tbl:uci_experiment_full} shows the mean test negative log predictive density (NLPD) over 10-fold cross-validation, where $\MZ, \vtheta, \valpha_{\vu}$, and $\vBeta_{\vu}$ are optimized. Other evaluation metrics, setup, and implementation details are available in \cref{appendix:uci}. In the streaming setting, the proposed method performs comparable to \citet{bui2017streaming}.
\paragraph{Sequential Learning of Magnetic Anomalies} We consider the task of online mapping of local anomalies in the ambient magnetic field. We follow the experiment setup and data provided in \citet{solin2018modeling}, where a small wheeled robot equipped with a 3-axis magnetometer moves around in a $\SI{6}{\meter} \times \SI{6}{\meter}$ indoor space. We assign a GP prior $\mathcal{GP}(0,\sigma_0^2+\kappa^\mathrm{Mat.}_{\sigma^2,\ell}(\vx,\vx'))$ to the magnetic field strength over the space under the presence of Gaussian measurement noise with variance $\sigma_\mathrm{n}^2$.

We include two experiments: 
In \cref{fig:robot-a}, we simultaneously learn the hyperparameters ($\sigma_0^2, \sigma^2, \ell, \sigma_\mathrm{n}^2$) and a representation in terms of inducing points and memory. Our method is able to form a representation by spreading inducing points and learning the hyperparameters progressively. This has practical importance in real-world robot estimation tasks, where the robot is not constrained to a predefined area \citep[a weakness in][]{solin2018modeling}.
In \cref{fig:robot-b}, we now receive data continuously during exploration. The visualization shows the mean estimate with marginal variance (uncertainty) controlling the opacity. We recover the same local estimate as in experiment \cref{fig:robot-a} and outperform the baseline given by \citet{bui2017streaming}. See details in \cref{appendix:magnetic}.\looseness-1

\subsection{Continual Learning on Split MNIST}
\label{sec:mnist}
\begin{figure}[b!]
	\raggedleft\footnotesize
	\setlength{\figurewidth}{.8\columnwidth}
	\setlength{\figureheight}{.66\figurewidth}
	\pgfplotsset{axis on top, scale only axis,ytick={0,.2,.4,.6,.8,1},yticklabels={0\%,20\%,40\%,60\%,80\%,100\%},xmin=0,xmax=20,ymin=-.3,ymax=1,xtick={0,4,8,12,16,20},xticklabel={~},y grid style={line width=.1pt, draw=gray!10,dashed},x grid style={solid},grid}
	\input{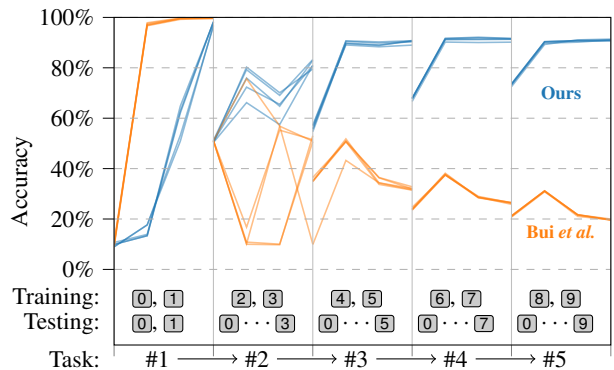}\\[-4em]
	\begin{tikzpicture}[outer sep=0pt]
		\draw[draw=none,fill=none] (0,0) rectangle ++({\figurewidth+3pt},-.5cm);
		
		\foreach \x/\y [count=\i] in {0/1,2/3,4/5,6/7,8/9}{
			\node[anchor=east,align=center,minimum width=0.2\figurewidth] at ({0.2*\figurewidth*\i},0) {\digit{\x}, \digit{\y}};
			\ifthenelse{\i=1}{\def\connector{, }}{\def\connector{$\,\cdots$}}
			\node[anchor=east,align=center,minimum width=0.2\figurewidth] at ({0.2*\figurewidth*\i},-.33cm) {\digit{0}\connector\digit{\y}};
			\node[anchor=center] (task-\i) at ({0.2*\figurewidth*\i-.1\figurewidth},-.8cm) {\#\i};
			
		}
		\node[anchor=east] at ({-0.00*\figurewidth},0) {Training:};
		\node[anchor=east] at ({-0.00*\figurewidth},-.33cm) {Testing:};
		\node[anchor=east] at ({-0.00*\figurewidth},-.8cm) {Task:}; 
		\draw[-latex',->] (task-1)--(task-2); 
		\draw[-latex',->] (task-2)--(task-3);     
		\draw[-latex',->] (task-3)--(task-4);
		\draw[-latex',->] (task-4)--(task-5);
	\end{tikzpicture}
	\vspace*{-6pt}
	\caption{Accuracy on {\em split MNIST} over training with different random seeds. The training starts with \protect\digit{0} vs.\ \protect\digit{1} and each task introduces new digits while testing on all classes thus far. The overall accuracy (mean over tasks) drops when introducing a new task, but recovers and does not suffer from forgetting over tasks.\looseness-1}
	\label{fig:split-mnist-acc}
\end{figure}

{\em Split MNIST} \citep{zenke17a} is a continual learning benchmark and a variant of MNIST where training data comes in five batches of two digits each, \ie \digit{0}~and~\digit{1}, \digit{2}~and~\digit{3}, \dots, \digit{8}~and~\digit{9}. Performance is measured by multi-class classification accuracy on all digits seen thus far. The model at each step has access only to the current batch of the classification task and thus should learn incrementally on different tasks without forgetting previous ones.  
\cref{fig:split-mnist-acc} compares the accuracy during the training on each task, subdivided into batches.
Our sequential model does well at remembering previous tasks (only marginal drops in accuracy in each column of \cref{tbl:split_mnist_acc}), while the model of \citet{bui2017streaming} forgets the previous tasks. We also slightly outperform \citet{kapoor2021variational}, for whom compute scales cubically with the number of tasks and who need an additional hyperparameter regularization term. Therefore, we distinguish it from other baselines which do not suffer from the same complexity with the number of tasks (\cref{tbl:split_mnist_methods_acc}).

\begin{table}[b!]
    \vspace*{-1em}
	\centering\footnotesize
	\caption{Test accuracy (and standard deviation over different random seeds) on {\em split MNIST} over all tasks thus far for the proposed method. High accuracy over all previous tasks shows that the method does not suffer from forgetting.}
	\label{tbl:split_mnist_acc}
	\setlength{\tabcolsep}{3pt}
	\begin{tabularx}{\columnwidth}{c c c c c c} 
		\toprule
		\textbf{Task} & \digit{0}, \digit{1} & \digit{2}, \digit{3} & \digit{4}, \digit{5} & \digit{6}, \digit{7} & \digit{8}, \digit{9}\\
		\midrule
		\#1	  &  $.98(.005)$ &  &  &  &  \\
		\#2   &  $.68(.029)$ & $.95(.003)$  &  &  & \\
		\#3   &  $.88(.005)$ & $.87(.004)$  & $.98(.002)$ & & \\
		\#4	  &  $.86(.006)$ & $.89(.004)$  & $.94(.007)$ & $.97(.004)$ & \\
		\#5	  &  $.95(.014)$ & $.87(.011)$ & $.90(.011)$ & $.93(.005)$ & $.90(.014)$\\
		\bottomrule
	\end{tabularx}
\end{table}

We include an experiment to show that our BLS is a useful metric to characterise and determine the difficulty of input points. We consider continual learning on {\em split MNIST} as before, but now we select data points from our training set and move them to the test set. Points are chosen either randomly or based on the BLS score. We then retrain the model on the reduced training set and test on the increased test set. \Cref{fig:mnist-bls-random} shows the performance of the selection methods.
Random selection of the points has a small negative effect on performance. However, using the BLS score is detrimental, showing the importance of the examples for the model that are moved to the test set. \Cref{fig:mnist-bls} shows digits with the highest BLS score. We also study the effect of the size of the memory, and show how it affects the model accuracy. We train our sequential model with different memory sizes using BLS and report test accuracy (see \cref{fig:split-mnist-mem-acc}). As expected, the accuracy increases with memory size, but remarkably a memory size of just 5\% achieves satisfactory performance. Experiment details can be found in \cref{appendix:mnist} and further evaluation of the BLS score on UCI in \cref{tbl:uci_experiment_bls_random}.

\begin{table}[t!]
	\centering\footnotesize
	\vspace*{-.5em}
	\caption{Test final accuracy and NLPD (standard deviation over different random seeds) of various methods on the \emph{split MNIST} task. We include \citet{kapoor2021variational} as a baseline.}
	\label{tbl:split_mnist_methods_acc}
	\begin{tabularx}{\columnwidth}{l c c}
		\toprule
		\bf Method & \bf Accuracy & \bf NLPD \\
		\midrule
		SVGP (Baseline) & $.962 (.001)$ & $\phantom{0}0.155 (0.002)$  \\
		\citet{bui2017streaming} & $.200 (.001)$ & $\phantom{0}2.150 (0.019)$ \\
		Ours (without memory) & $.208 (.007)$& $68.038 (2.830)$\\		
		Ours (with memory) & $.909 (.001)$ & $\phantom{0}0.316 (0.010)$ \\
		\midrule
		\citet{kapoor2021variational} & $.905 (.010)$ & $\phantom{0}0.324 (0.018)$ \\
		\bottomrule
	\end{tabularx}
\end{table}

\subsection{Timing Experiment}
\begin{figure}[t!]
	\scriptsize
	\centering
	\setlength{\figurewidth}{.9\columnwidth}
	\setlength{\figureheight}{.4\figurewidth}
	\pgfplotsset{axis on top, scale only axis,legend style={font=\tiny},grid style={line width=.1pt, draw=gray!10,dashed},grid,y tick label style={rotate=90,font=\tiny},ylabel style={yshift=-2.5em},x tick label style={font=\tiny}}
	\input{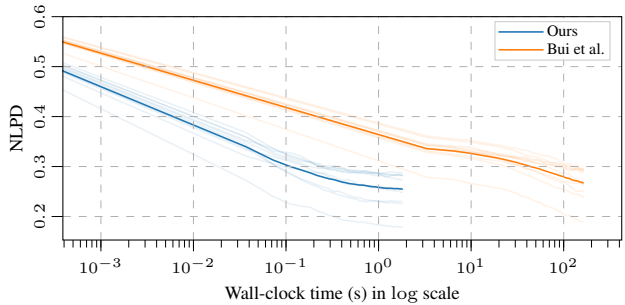}
	\vspace*{-2em}
	\caption{NLPD against wall-clock time of the proposed method and \citet{bui2017streaming} on `adult' UCI data set when only variational parameters are optimized. At the same performance, the proposed method is significantly faster than \citet{bui2017streaming}.}
	\label{fig:uci_wall_clock}
\end{figure}

Finally, to show the benefit of natural gradient optimization for variational parameters, we perform a timing experiment against \citet{bui2017streaming} in \cref{fig:uci_wall_clock} in a streaming setting on the `adult' UCI data set. We see speed improvements in inference wall-clock timings with the same performance (note the x-axis is in log-scale).

\section{Discussion and Conclusions}
The difficulty in building a sequential sparse GP model is due to the lack of access to previous data. Approaches that do not consider the dual parameter perspective fail to see the essence of the inference problem; how to accurately infer the dual parameters in a sequential manner. The problem formulation was presented in \citet{csato2002sparse}, but there an EP inference scheme was used. This paper shows how to update parameters sequentially using variational approximate inference. 

Furthermore, we use natural gradient updates which come easily in the dual parameter formulation allowing for a method (\ie, few inference step iterations) that works with general likelihoods. Lack of access to past data also makes it challenging to learn hyperparameters $\vtheta$. The problem worsens in the non-stationary continual learning setting. Here, to counter the problem of forgetting previous data, we introduce the concept of memory, a technique shown to have success in deep learning \citep{pan2020continual}. We find that a small amount of memory can dramatically improve performance by replicating the offline ELBO solution.

Our memory approach is novel and different from previous work, which attempts to replace missing data with additional regularization terms in the ELBO \citep{bui2017streaming, kapoor2021variational}. Our final method manages to control the error that come from inference, learning and representation (memory and inducing points). Furthermore, our method can be applied to a variety of sequential learning tasks.
Given the importance of memory to our method, a good selection technique is paramount. We derive a new Bayesian leverage score, given its connection to Bayesian approximate inference. The score generalizes the prevalent RLS score used for kernel sampling methods \citep{alaoui2015fast}. We find the importance of the dual parameters again, as they are a crucial component of the BLS. We demonstrate the applicability of our method to a complex combined batch Bayesian optimization and active learning problem, and continual learning problem. \looseness-1

A reference implementation of the methods presented in this paper is available at:
\url{https://github.com/AaltoML/sequential-gp}.

\section*{Acknowledgements}
This work was supported by funding from JST CREST (Grant ID JPMJ CR2112), the Academy of Finland (grant id 339730 and 324345), and the Finnish Center for Artificial Intelligence (FCAI). We acknowledge the computational resources provided by the Aalto Science-IT project and CSC -- IT Center for Science, Finland. We thank Rui Li, Riccardo Mereu, Henry Moss, Thomas M{\"o}llenhof, Gianma Marconi, and Victor Picheny for helpful discussions. 
We also acknowldge the input from Dharmesh Tailor, Siddharth Swaroop, and Eric Nalisnick related the BLS score derivation.\looseness-1

\bibliographystyle{icml2023}

\clearpage

\appendix

\onecolumn

\section*{Appendix}

We include technical details of the methods that were omitted for brevity in the main paper. Additionally, we provide details on the experiments and evaluation setup for reproducing the results in the main paper, and include further results, figures, and tables  that extend the evaluation. %

\section{Method Details}

\subsection{Equivalence to Csat\'o and Opper}
\label{appendix:opper_equiv}
The result from \citet{csato2002sparse} (\cf\ Lemma~1) assumes a Gaussian prior $p(\vf)$ with a mean function $m(\vx)$ and covariance $\kappa(\vx,\vx')$, with likelihood $p(\data\mid\vf)$ where $\vf = [ f_1, f_2, \ldots, f_n ]$ is a vector of function values $f_i = f(\vx_i)$.
For this case, they use the following two parameterization,
\begin{equation}
\begin{split}
   q_i &= \int \nabla_{f_i} p(\data \mid \vf) \frac{p(\vf)}{\int p(\data\mid\vf) p(\vf)\diff\vf} \diff\vf ,\\
   R_{i,j} &= \int \, \nabla^2_{f_i, f_j} p(\data \mid \vf) \frac{p(\vf)}{\int p(\data\mid\vf) p(\vf)\diff\vf} \diff\vf  - q_i q_j.
\end{split}
\end{equation}

For the case when $p(\data \mid \vf) = \prod_{i=1}^N p(y_i \mid f_i)$, we can show the following,
\[ q_i = \alpha_i \text{ and } R_{i,j} = \left\{ \begin{array}{ll} \beta_i, & \text{ when } i=j \\ 0, & \text{ when } i\ne j \end{array} \right. ,\]
where $\alpha_i$ and $\beta_i$ are as defined in \cref{eq:dual_param}. We first show the proof for $q_i$,
\begin{align}
   q_i &= \int \nabla_{f_i} p(\data \mid \vf) \frac{1}{p(\data\mid\vf)}  \frac{p(\data\mid\vf)p(\vf)}{\int p(\data\mid\vf) p(\vf)\diff\vf} \diff\vf \nonumber\\
   &= \int \nabla_{f_i} \log p(\data \mid \vf) p(\vf\mid\data) \diff\vf \nonumber\\
   &= \sum_{j=1}^N \int \nabla_{f_i} \log p(y_j \mid f_j) p(\vf\mid\data) \diff\vf \nonumber\\
   &= \int \nabla_{f_i} \log p(y_i \mid f_i) p(\vf\mid\data) \diff\vf \nonumber\\
   &= \int \nabla_{f_i} \log p(y_i \mid f_i) p(f_i\mid\data) \diff f_i \nonumber\\
   &= \myexpect_{p(f_i \mid \vy_n)} [\nabla_{f_i} \log p(y_i \mid f_i)].
\end{align}
Here, the first line is obtained by simply multiplying and dividing by $p(\data\mid\vf)$, and we get the second line by using the definition of the posterior $p(\vf\mid\data)$ and the fact that $\nabla \log p(\data\mid\vf) = [\nabla p(\data\mid\vf)]/p(\data\mid\vf)$. The third line follows from the assumption that the likelihood factorizes over data examples, and the fourth line obtained by noting that $\nabla_{f_i} \log p(y_j \mid f_j)$ is non-zero only when $i=j$. 
The fifth line follows by marginalizing out all $f_j$ other than $f_i$, and the final line is just a different way to write the expectation.

For $R_{i,j}$, we proceed in a similar fashion. We will use the following identity to write $\nabla_{f_i, f_j} p(\data \mid \vf)$ in terms $\nabla_{f_i} \log p(\data\mid\vf)$:
\begin{equation}
   \frac{1}{p(\data\mid\vf)} \nabla^2_{f_i f_j} p(\data\mid\vf) = \nabla^2_{f_i f_j} \log p(\data\mid\vf) + \sqr{ \nabla_{f_i} \log p(\data\mid\vf) } \sqr{\nabla_{f_j} \log p(\data\mid\vf)}.
   \label{eq:identify_1}
\end{equation}
This can be proved by rearranging the derivative of $\nabla^2_{f_i f_j} \log p(\data\mid\vf)$ by using the fact that $\nabla \log p(\data\mid\vf) = [\nabla p(\data\mid\vf)]/p(\data\mid\vf)$ (which we also used in the derivation of $q_i$ above). Using this we can simplify $R_{i,j}$,
\begin{align}
   R_{i,j} &= \int \, \nabla^2_{f_i, f_j} p(\data \mid \vf) \frac{1}{p(\data\mid\vf)} \frac{ p(\data\mid\vf) p(\vf)}{\int p(\data\mid\vf) p(\vf)\diff\vf} \diff\vf  - q_i q_j \nonumber \\
   &= \int \, \sqr{ \nabla^2_{f_i f_j} \log p(\data\mid\vf) + \sqr{ \nabla_{f_i} \log p(\data\mid\vf) } \sqr{\nabla_{f_j} \log p(\data\mid\vf)} } p(\vf\mid\data) \diff\vf  - q_i q_j \nonumber \\
   &= \int \, \sqr{ \nabla^2_{f_i f_j} \log p(\data\mid\vf)} p(\vf\mid\data) \diff\vf + \int \sqr{ \nabla_{f_i} \log p(\data\mid\vf) } \sqr{\nabla_{f_j} \log p(\data\mid\vf)}  p(\vf\mid\data) \diff\vf  - q_i q_j \nonumber\\
   &= \sum_{k=1}^n \int \, \sqr{ \nabla^2_{f_i f_j} \log p(y_k \mid f_k)} p(\vf\mid\data) \diff\vf + \sum_{l,m} \int \sqr{ \nabla_{f_i} \log p(y_l\mid f_l) } \sqr{\nabla_{f_j} \log p(y_m\mid f_m)}  p(\vf \mid\data) \diff\vf  - q_i q_j, 
\end{align}
where in the first line we multiply and divide by $p(\data\mid\vf)$ and use \cref{eq:identify_1} and definition of the posterior to get to the second line. The third line is a rearrangement, and the last line is obtained by using the factorial property of the likelihood.

For the second term we get it to be non-zero when $i=l$ and $j=m$, and then integrating over all function values, except $f_i$ and $f_j$, the term reduces to $q_i q_j$, which can be cancelled out. This gives us the following expression,
\begin{equation}
\begin{split}
   R_{i,j} &= \sum_{k=1}^n \int \, \sqr{ \nabla^2_{f_i f_j} \log p(y_k \mid f_k)} p(\vf\mid\data) \diff\vf   \\ 
   &= \left\{ \begin{array}{ll} \myexpect_{p(f_i \mid \vy_n)} [\nabla_{f_i f_j}^2 \log p(y_i \mid f_i)], & \text{ when } i=j, \\ 0, & \text{ when } i\ne j. \end{array} \right.
\end{split}
\end{equation}
This is because the derivative inside is only nonzero when $i=j=k$, and we can express the integral only over $f_i$. This proves the required result.

\subsection{Derivation of the Dual-form of the SVGP Stationary Point}\label{app:dual_params}
We will derive the dual-form of the stationary point of the following SVGP objective:
\begin{equation}\label{eq:elbo_svgp_}
  \mathcal{L}(q) = \sum_{\mathclap{i \in \mathcal{D}}} \mathbb{E}_{q_{\vu}(f_i)} [\log p(y_i \mid f_i)]- \dkl{}{q_{\vu}(\vu)}{p_{\vtheta}{(\vu)}}.
\end{equation} 
Derivation simplifies by using natural and expectation parameters of $q_{\vu}(\vu) = \N(\vu|\vm,\MV)$, denoted by $\vlambda$ and $\vmu$ respectively,
\[
   \vlambda = (\vlambda^{(1)}, \vlambda^{(2)}) = (\vV^{-1} \vm, \,\, -\frac{1}{2} \vV^{-1}), \qquad
   \vmu = (\vmu^{(1)}, \vmu^{(2)}) = (\vm, \,\, \vm\vm^\top + \vV),
\]
Taking the gradient with respect to $\vmu$ of \cref{eq:elbo_svgp_} at a stationary point $q_{\vmu}^*(\vu)$, we get
 \begin{align}
 \vzeros  &= \sum_{\mathclap{i \in \mathcal{D}}} \nabla_{\vmu}  \mathbb{E}_{q^*_{\vu}(f_i)} [\log p(y_i \mid f_i)] - \nabla_{\vmu}\dkl{}{q^*_{\vu}(\vu)}{p_{\vtheta}{(\vu)}}, \nonumber\\
    &=  \sum_{\mathclap{i \in \mathcal{D}}} \nabla_{\vmu}  \mathbb{E}_{q^*_{\vu}(f_i)} [\log p(y_i \mid f_i)] - (\vlambda_* - \vlambda_\text{prior}), \nonumber\\
  &\implies \vlambda_* = \vlambda_\text{prior} + \sum_{\mathclap{i \in \mathcal{D}}} \nabla_{\vmu}  \mathbb{E}_{q^*_{\vu}(f_i)} [\log p(y_i \mid f_i)], \label{eq:fixed point}
 \end{align}
 where $\vlambda_\text{prior} = (0, -\MKzz^{-1}/2)$ is the natural parameter of the prior $p_{\vparam}(\vu)$.

We can expand this equation and write two equations corresponding to the two natural parameters $(\MV_*^{-1} \vm_*, \, -\MV_*^{-1}/2)$. We get the following for the first natural parameter,
 \begin{align}
    \MV^{-1}_*\vm_* &= \sum_{\mathclap{i \in \mathcal{D}}} \nabla_{\vmu^{(1)}}  \mathbb{E}_{q^*_{\vu}(f_i)} [\log p(y_i \mid f_i)] \nonumber \\
    &= \sum_{\mathclap{i \in \mathcal{D}}}  \va_i \mathbb{E}_{q^*_{\vu}(f_i)} [\nabla \log p(y_i \mid f_i)] - \sum_{\mathclap{i \in \mathcal{D}}} \va_i \mathbb{E}_{q^*_{\vu}(f_i)} [ \nabla^2 \log p(y_i \mid f_i)] \va_i^\top \vm_* \nonumber \\
    &= \sum_{\mathclap{i \in \mathcal{D}}}  \va_i ( \hat{\alpha}_i^* + \hat{\beta}_i^* \va_i^\top \vm_* )  \label{eq:first_param}
 \end{align}
 In the second line we used the identity given in \citet[Eq. 10]{khan2021BLR} which uses chain-rule and Bonnet's result. This gives us \cref{eq:nat1}. This also matches with Eq. 21 of \citet{adam2021dual} who define $\lambda_{1,i}^* = \hat{\alpha}_i^* + \hat{\beta}_i^* \va_i^\top \vm_* $.

 Similarly for the second natural parameter,
 \begin{align}
 -\frac{1}{2} \MV_*^{-1} &= -\frac{1}{2} \MKzz^{-1} + \sum_{\mathclap{i \in \mathcal{D}}} \nabla_{\vmu^{(2)}}  \mathbb{E}_{q^*_{\vu}(f_i)} [\log p(y_i \mid f_i)] \nonumber \\
    &= -\frac{1}{2} \MKzz^{-1} +  \half \sum_{\mathclap{i \in \mathcal{D}}} \va_i \mathbb{E}_{q^*_{\vu}(f_i)} [ \nabla^2 \log p(y_i \mid f_i)] \va_i^\top \nonumber \\
    &= -\frac{1}{2} \MKzz^{-1} -  \half \sum_{\mathclap{i \in \mathcal{D}}} \va_i \hat{\beta}_i^* \va_i^\top \label{eq:second_param}
 \end{align}
Here again, in the second line we used the identity given in \citet[Eq. 11]{khan2021BLR} which uses chain-rule and Price's result. This gives us \cref{eq:nat2}. This too matches with Eq. 21 of \citet{adam2021dual}.

We will now derive the dual-form similar to \cref{eq:gp_pred}. Rearranging and substituting \cref{eq:second_param} in to \cref{eq:first_param}, we get,
 \begin{align}
     [\MV_*^{-1} - \va_i\hat{\beta}_i^* \va_i^\top]\vm_* = \textstyle\sum_{{i \in \mathcal{D}}} \va_i \hat{\alpha}_i^* \qquad 
    &\implies \MKzz^{-1} \vm_* =  \MKzz^{-1}   \underbrace{ \textstyle\sum_{{i \in \mathcal{D}}} \vkzi  \hat{\alpha}_i^* }_{ = \valpha_{\vu}} \nonumber\\
     \MV_*^{-1} = \MKzz^{-1} + \MKzz^{-1}  ( \textstyle\sum_{{i \in \mathcal{D}}} \vkzi  \hat{\beta}_i^* \vkzi^\T ) \MKzz^{-1} \qquad
   &\implies \MV_*^{-1} = \MKzz^{-1} +  \MKzz^{-1} \underbrace{\textstyle\sum_{{i \in \mathcal{D}}} \vkzi  \hat{\beta}_i \vkzi^\T}_{ = \vBeta_{\vu}}    \MKzz^{-1} 
    \label{eq:rearrang_dual}
 \end{align}
Now substituting these results into \cref{eq:prediction}, we get the results:
\begin{align}
\begin{split}
\myexpect_{q_{\vu}^*(f)}[f_i] &=  \vkzs^{\T} \MKzz^{-1} \vm^* = \vkzs^{\T} \MKzz^{-1} \valpha_{\vu}, \\
\textrm{Var}_{q_{\vu}^*(f)}[f_i]  &=  \kappa_{ii} - \vkzs^\top \sqr{\MKzz^{-1} - \MKzz^{-1}  \MV_* \MKzz^{-1}}   \vkzs =  \kappa_{ii} - \vkzs^\top \sqr{\MKzz^{-1} - \rnd{ \MKzz + \vBeta_{\vu}^* }^{-1} }   \vkzs
\end{split}
\end{align}
Thus proving that we can rewrite the posterior process in terms of the sparse dual parameters.

\subsection{Derivation of the Pseudo-Data and New Prior}\label{app:MVN_correlated}
We first derive the distribution form used in \cref{eq:prior_alt} for the following,
\begin{equation} 
   \hat{p}_{\data}(\vu) \propto \prod_{i\in \mathcal{D}} e^{ {-\half \hat{\beta}_i^* (\hat{y}_i^* - \va_i^\top \vu)^2} }= \N( \vu \mid \tilde{\vy}^{*}, \tilde{\vSigma}^{*}),
\end{equation}
where $(\hat{\beta}_i^*, \hat{y}_i^*)$ are obtained by training on the data $\data$. We start by writing the product in a matrix form,
\begin{equation} 
 \begin{split}
    \log \hat{p}_{\data}(\vu) &= - \half \vu^\top (\tilde{\vSigma}^*)^{-1} \vu + \vu^\top (\tilde{\vSigma}^*)^{-1} \tilde{\vy} + \text{const.}\\
    \log \prod_{i\in \mathcal{D}} e^{ {-\half \hat{\beta}_i^* (\hat{y}_i^* - \va_i^\top \vu)^2} } &=  -\half \vu^\top \MA^\top  \diag(\hat{\vbeta}^*) \MA \vu + \vu^\top \MA^\top \diag(\hat{\vbeta}^*) \hat{\vy}^* +\text{const.}%
 \end{split}
\end{equation}
where $\MA = \MKxz\MKzz^{-1}$, and $\hat{\vy}^*$ and $\hat{\vbeta}^*$ are vectors of all $\hat{y}_i^*$ and $\hat{\beta}_i^*$ respectively. Then, we simply match the terms in $\vu$. First, by matching the quadratic term, we get, 
\begin{align}
   (\tilde{\vSigma}^*)^{-1} = \MA^\T \diag(\hat{\vbeta}^*)\MA
   = \MKzz^{-1}\MKzx \diag(\hat{\vbeta}^*) \MKxz \MKzz^{-1} 
   = \MKzz^{-1} \vBeta_{\vu}^* \MKzz^{-1} 
\end{align}
which gives us the third equation in \cref{eq:phat}. Next, by matching the linear term in $\vu$, we get,
\begin{align}
\tilde{\vy}^* &= \tilde{\vSigma}^* \MA^\T \diag(\hat{\vbeta}^*)\hat{\vy}^* \nonumber \\
 &= \MKzz [\vBeta^*_{\vu}]^{-1} \MKzz \MKzz^{-1}\MKzx \diag(\hat{\vbeta}^*)\hat{\vy}^* \nonumber \\
 &= \MKzz [\vBeta^*_{\vu}]^{-1} \MKzx \diag(\hat{\vbeta}^*)[ \diag(\hat{\vbeta}^*)^{-1} \hat{\valpha}^* + \MA\vm^* ]  \nonumber \\
 &= \MKzz [\vBeta^*_{\vu}]^{-1} \MKzx \hat{\valpha}^* + \MKzz [\vBeta^*_{\vu}]^{-1}   \MKzx \diag(\hat{\vbeta}^*) \MKxz \MKzz^{-1} \vm^* \nonumber\\
  &= \MKzz [\vBeta^*_{\vu}]^{-1} \valpha^*_{\vu} + \MKzz [\vBeta^*_{\vu}]^{-1}  \vBeta^*_{\vu}  \MKzz^{-1} \vm^*  \nonumber\\
  &= \MKzz [\vBeta^*_{\vu}]^{-1} \valpha^*_{\vu} + \vm^*,
\end{align}
which recovers the second equation in \cref{eq:phat}.
Note that we can use different forms as well. For example, an equivalent way to write \cref{eq:phat} is
\begin{align}
   \hat{p}_{\data_\text{old}}(\vu) &= \N( \vu | \tilde{\vSigma}^\text{old} (\MV^\text{old})^{-1} \vm^\text{old}, \, \,  \tilde{\vSigma}^\text{old}) , \\
   \tilde{\vy}^\text{old} &=  \tilde{\vSigma}^\text{old} (\MV^\text{old})^{-1} \vm^\text{old}, \\
   \tilde{\vSigma}^\text{old} &= \sqr{ (\MV^\text{old})^{-1} - \MKzz^{-1} }^{-1}, \nonumber 
\end{align}
which is in terms of the mean and covariance of the posterior.
This expression is similar to the one used in \citet[Eq. 7][]{bui2017streaming}. All such forms are equivalent. We use \cref{eq:phat} because it uses dual parameters which are readily obtained through the natural-gradient method of \citet{adam2021dual}.

We can write $\hat{p}_{\mathcal{M}}(\vu)$ in a similar way,
\begin{align}
   \hat{p}_{\mathcal{M}}(\vu) &\propto \prod_{i\in \mathcal{M}} e^{ {-\half \hat{\beta}_i^\text{old} (\hat{y}_i^\text{old} - \va_i^\top \vu)^2} } = \N( \vu \mid \tilde{\vy}^{\mathcal{M}}, \tilde{\vSigma}^{\mathcal{M}}).
\end{align}
The expressions for $\tilde{\vy}^{\mathcal{M}}$ and $\tilde{\vSigma}^{\mathcal{M}}$ are derived by repeating the same derivation but now only involving $i \in \mathcal{M}$.
For example,
\begin{align}
   (\tilde{\vSigma}^\mathcal{M})^{-1} &= \sum_{i\in\mathcal{M}} \va_i \hat{\beta}_i^\text{old}\va_i^\top 
   = \MKzz^{-1} \underbrace{ \rnd{ \textstyle \sum_{i\in\mathcal{M} } \vkzi \hat{\beta}_i^{\text{old}} \vkzi^\top } }_{= \vBeta_{\vu}^{\mathcal{M}}} \MKzz^{-1}  \nonumber\\
   \tilde{\vy}^\mathcal{M} &= \tilde{\vSigma}^{\mathcal{M}} \sum_{i\in \mathcal{M}} \va_i \hat{\beta}_i^\text{old} \hat{y}_i^\text{old} 
   = \MKzz [\vBeta^{\mathcal{M}}_{\vu}]^{-1} \sum_{i\in \mathcal{M}} \vkzi (\hat{\alpha}_i^\text{old} + \hat{\beta}_i^\text{old} \va_i^\top \vm^\text{old}) \nonumber\\
   &= \MKzz [\vBeta^{\mathcal{M}}_{\vu}]^{-1} \underbrace{ \sum_{i\in \mathcal{M}} \vkzi \hat{\alpha}_i^\text{old} }_{= \valpha_{\vu}^{\mathcal{M}}}  + \sqr{ \MKzz [\vBeta^{\mathcal{M}}_{\vu}]^{-1} \bigg( \sum_{i\in\mathcal{M}} \vkzi \hat{\beta}_i^\text{old} \vkzi \bigg) \MKzz^{-1} } \vm^* \nonumber\\
   &= \MKzz [\vBeta^{\mathcal{M}}_{\vu}]^{-1} \valpha_{\vu}^{\mathcal{M}}  + \vm^* .
\end{align}
Next, we derive the mean and covariance of the new prior $ \hat{q}_{\vu}^\text{old}(\vu) /  \hat{p}_{\mathcal{M}}(\vu) \propto \N(\vu| \vm_{\text{prior}}, \MV_{\text{prior}})$. Let us denote the current estimate of the dual pair by $(\valpha_{\vu}^\text{old}, \vBeta_{\vu}^\text{old})$ as defined in \cref{eq:dual_sparse}. There, the whole $\mathcal{D}_\text{old}$ is used but, in reality, the dual pairs are learned sequentially and may not represent the exact dual parameters. Still, we use the same notation for
convenience.
The current natural-parameters of the posterior can be then written by rewriting \cref{eq:nat1,eq:nat2} in terms of the dual pairs,
\begin{equation}
\begin{split}
   (\hat{\MV}^{\text{old}})^{-1} \hat{\vm}^{\text{old}} &= \MKzz^{-1} \valpha_{\vu}^\text{old} + \MKzz^{-1} \vBeta_{\vu}^\text{old} \MKzz^{-1} \vm^\text{old}, \\
   (\hat{\MV}^{\text{old}})^{-1} &= \MKzz^{-1} \vBeta_{\vu}^\text{old} \MKzz^{-1} + \MKzz^{-1} .
\end{split}
\end{equation}
Similarly, we can write the natural parameters of $\hat{p}_{\mathcal{M}}(\vu)$ in terms (of the current estimate) of $(\valpha_{\vu}^{\mathcal{M}}, \vBeta_{\vu}^{\mathcal{M}})$,
\begin{equation}
\begin{split}
   (\hat{\MV}^{\mathcal{M}})^{-1} \hat{\vm}^{\mathcal{M}} &= (\tilde{\vSigma}^{\mathcal{M}})^{-1} \tilde{\vy}^{\mathcal{M}} = \MKzz^{-1} \valpha_{\vu}^{\mathcal{M}}  + \MKzz^{-1}\vBeta_{\vu}^{\mathcal{M}} \MKzz^{-1} \vm^* ,\\
   (\hat{\MV}^{\mathcal{M}})^{-1} &= (\tilde{\vSigma}^{\mathcal{M}})^{-1} = \MKzz^{-1}\vBeta_{\vu}^{\mathcal{M}} \MKzz^{-1}. 
\end{split}
   \label{eq:natparam_old}
\end{equation}
The natural parameters of the new prior are simply obtained by subtracting the natural parameters given above,
\begin{equation}
\begin{split}
   (\MV^{\text{prior}})^{-1} \vm^{\text{prior}} &= (\hat{\MV}^{\text{old}})^{-1} \hat{\vm}^{\text{old}} - (\hat{\MV}^{\mathcal{M}})^{-1} \hat{\vm}^{\mathcal{M}} = \MKzz^{-1} \valpha_{\vu}^{\text{old} \backslash \mathcal{M}} + \MKzz^{-1} \vBeta_{\vu}^{\text{old} \backslash \mathcal{M}} \MKzz^{-1} \vm^\text{old} , \\
   (\MV^{\text{prior}})^{-1} &= (\hat{\MV}^{\text{old}})^{-1} - (\hat{\MV}^{\mathcal{M}})^{-1} = \MKzz^{-1} \vBeta_{\vu}^{\text{old} \backslash \mathcal{M}} \MKzz^{-1} + \MKzz^{-1} , 
   \label{eq:nat_param_prior} 
\end{split}
\end{equation}
where $(\valpha_{\vu}^{\text{old} \backslash \mathcal{M}}, \, \vBeta_{\vu}^{\text{old} \backslash \mathcal{M}})$ are defined in \cref{eq:nat1_M,eq:nat2_M}.

\subsection{Derivation of the Natural-Gradient Descent Algorithm}
\label{app:ngd_deriv}
We will optimize the new objective in \cref{eq:elbo_new1} by using the Bayesian learning rule (BLR) of \citet{khan2021BLR} which is a natural-gradient descent algorithm. We start by denoting the natural and expectation parameters of a posterior $q_{\vu}^{(t)}(\vu)$ obtained in the $t$'th iteration by $\vlambda^{(t)}$ and $\vmu^{(t)}$ respectively. We denote the natural parameters of the prior $ \hat{q}_{\vu}^\text{old}(\vu) /  \hat{p}_{\mathcal{M}}(\vu) \propto \N(\vu| \vm_{\text{prior}},
\MV_{\text{prior}})$ by $\vlambda_\text{prior}$; an expression is given in \cref{eq:nat_param_prior}. 
With these, the BLR update can be written as the following,
\begin{equation}
   \vlambda^{(t)} \leftarrow (1-\rho) \vlambda^{(t-1)} + \rho \bigg( \sum_{i\in\data_\text{new}} \left. \nabla_{\vmu} \,   \mathbb{E}_{q_{\vu}(f_i)} [\log p(y_i \mid f_i)] \right\vert_{\vmu=\vmu^{(t)}} + \vlambda_{\text{prior}} \bigg).
\end{equation}
To simplify the implementation, we will write the updates in terms of the estimate of the dual pair $(\valpha_{\vu}^{(t)}, \vBeta_{\vu}^{(t)})$ at iteration $t$. We make use of the fact that each iterate $\vlambda^{(t)}$ has the same dual form as in \cref{eq:natparam_old}. This is written below,
\begin{equation}
\begin{split}
   (\MV^{(t)})^{-1} \vm^{(t)} &= \MKzz^{-1} \valpha_{\vu}^{(t)} + \MKzz^{-1} \vBeta_{\vu}^{(t)}\MKzz^{-1} \vm^{(t)} ,\\
   (\MV^{(t)})^{-1} &= \MKzz^{-1} \vBeta_{\vu}^{(t)} \MKzz^{-1} + \MKzz^{-1}.
\end{split}
   \label{eq:natparam_itert}
\end{equation}
As shown in \cref{eq:nat_param_prior}, the prior $\vlambda_\text{prior}$ too has the same form written in terms of the dual parameters $(\valpha_{\vu}^{\text{old} \backslash \mathcal{M}}, \, \vBeta_{\vu}^{\text{old} \backslash \mathcal{M}})$. Finally, as shown in \cref{eq:first_param,eq:second_param}, the natural-gradients too can be written in the same form,
\begin{align}
   \textstyle\sum_{i\in\data_\text{new}} \left. \nabla_{\vmu^{(1)}}  \mathbb{E}_{q_{\vu}(f_i)} [\log p(y_i \mid f_i)] \right\vert_{\vmu=\vmu_t} &= \MKzz^{-1} \rnd{ \textstyle\sum_{i\in\data_\text{new}} \vkzi \hat{\alpha}_i^{(t)} } + \MKzz^{-1} \rnd{ \textstyle\sum_{i\in\data_\text{new}} \vkzi \hat{\beta}_i^{(t)} \vkzi^\top } \MKzz^{-1} \vm^{(t)}, \\
   \textstyle\sum_{i\in\data_\text{new}} \left. \nabla_{\vmu^{(2)}}  \mathbb{E}_{q_{\vu}(f_i)} [\log p(y_i \mid f_i)] \right\vert_{\vmu=\vmu_t} &= \MKzz^{-1} \rnd{ \textstyle\sum_{i\in\data_\text{new}} \vkzi \hat{\beta}_i^{(t)} \vkzi^\top } \MKzz^{-1} ,%
\end{align}
where $\hat{\alpha}_{i}^{(t)}$ and $\hat{\beta}_{i}^{(t)}$ are defined similarly to \cref{eq:nat1,eq:nat2}, but now by using $q_{\vu}^{(t-1)}(f_i)$,
\begin{equation}
\begin{aligned}
   \hat{\alpha}_i^{(t)} &= \myexpect_{q^{(t-1)}_{\vu}(f_i)} [\nabla_{f_i}\log p(y_i \mid f_i)], \qquad
   \hat{\beta}_i^{(t)} = \myexpect_{q^{(t-1)}_{\vu}(f_i)} [ -\nabla^2_{f_i}\log p(y_i \mid f_i)].
\end{aligned}
\end{equation}
We can use these to simply write the update in terms of the dual pair. Essentially, we use the following equivalent update,
\begin{equation}
    \begin{aligned}
       \valpha_\vu^{(t)} &\leftarrow (1 -\rho)\valpha_\vu^{(t-1)}  + \rho \rnd{ \valpha_{\vu}^{\text{old} \backslash \mathcal{M}} + \textstyle\sum_{i\in\data_\text{new}} \vkzi \hat{\alpha}_i^{(t)} } , \\
       \vBeta_{\vu}^{(t)} &\leftarrow (1 - \rho)\vBeta_{\vu}^{(t-1)} + \rho \rnd{ \vBeta_{\vu}^{\text{old} \backslash \mathcal{M}} + \textstyle\sum_{i\in\data_\text{new}} \vkzi \hat{\beta}_i^{(t)} \vkzi^\top}.
\end{aligned}
\end{equation}
This is followed by an update of the mean and the covariance given below,
\begin{equation}
\begin{split}
   \vm^{(t)} &\leftarrow \valpha_{\vu}^{(t)}, \qquad\qquad
   \MV^{(t)} \leftarrow \rnd{ \MKzz^{-1} \vBeta_{\vu}^{(t)} \MKzz^{-1} + \MKzz^{-1} }^{-1}.
\end{split}
   \label{eq:meancov_itert}
\end{equation}
This is derived by using \cref{eq:natparam_itert} and simplifying similarly to the first equation in \cref{eq:rearrang_dual}.

\subsection{Bayesian Optimization / Active Learning Algorithm}
\label{app:BO}
In \cref{alg:BO_fast_conditioning}, we include the algorithm that is used in the Bayesian optimization experiment in the main paper (\cref{sec:BO_AL}), where we fantasize a batch with dual conditioning. The algorithm uses the method outlined in the paper combined with any simple acquisition function $\gamma(\cdot)$.

\begin{algorithm}[h!]
	\caption{Fantasizing a batch with Dual Conditioning.} \label{alg:BO_fast_conditioning}
	\textbf{Input:} current model parameters $\vtheta$, $\MZ$, $(\valpha_{\vu}, \vBeta_{\vu})$, acquisition function $\gamma(\cdot)$, batch size $k$ \\
	\textbf{Initialize:} $\MX_b = \emptyset$
	\begin{algorithmic}[1] 
		\For{$i$ in $1,2,\ldots,k$}\Comment{$k$ is desired number of query points}
		\State $\vx_i = \mathrm{arg}\,\max_{\vx} \gamma(\vx)$ \Comment{Calculate $\gamma(\vx)$ using prediction function \cref{eq:prediction} at $\vx$}
		\State $y_i = \mathbb{E}[f(\vx_i)]$ \Comment{Fantasized $y$ is mean of the GP at $\vx_i$}
		\State $\mathcal{D}_{\textrm{new}} = (\vx_i,y_i)$ \Comment{The fantasized data point is treated as new data}
		\State Compute $(\valpha_{\vu}, \vBeta_{\vu})$ using $\mathcal{D}_{\textrm{new}}$ using method in \citet{adam2021dual} \Comment{Dual conditioning}
		\State $\MX_{\textrm{b}} \leftarrow \MX_{\textrm{b}} \cup \vx_i$ \Comment{$\vx_i$ is added to the current batch points}
		\EndFor
	\end{algorithmic}
	\textbf{Return:} $\MX_{\textrm{b}}$ \Comment{$\MX_{\textrm{b}}$ is the chosen batch of points.}
\end{algorithm}

\clearpage

\section{Experiment Details}
\label{app:experiments}
In \cref{sec:experiments}, we performed a series of experiments and ablation studies to showcase the capability of our proposed method in various setups. We also compared against other methods, in particular \citet{bui2017streaming} and \citet{maddox2021conditioning}. Here, we provide further details regarding the setup and the experiments performed.

\subsection{Streaming Banana Data Set}
\label{appendix:banana}
The streaming banana classification experiment was used by both \citet{bui2017streaming} and \citet{maddox2021conditioning}. The data set is divided into four batches of $100$ points each. First, we compare within the setup of \citet{maddox2021conditioning}, who focused on fast conditioning, but kept the hyperparameters fixed. Second, we compare with \citet{bui2017streaming}, who address hyperparameter learning without considering the speed of conditioning.

\begin{figure*}[t!]
	\centering
	\setlength{\figurewidth}{.3\textwidth}
	\setlength{\figureheight}{\figurewidth}
	\pgfplotsset{scale only axis,axis on top}
	\begin{subfigure}[b]{.3\textwidth}
		\centering
		\begin{tikzpicture}[inner sep=0, outer sep=0]
			\node{\includegraphics[width=\figurewidth]{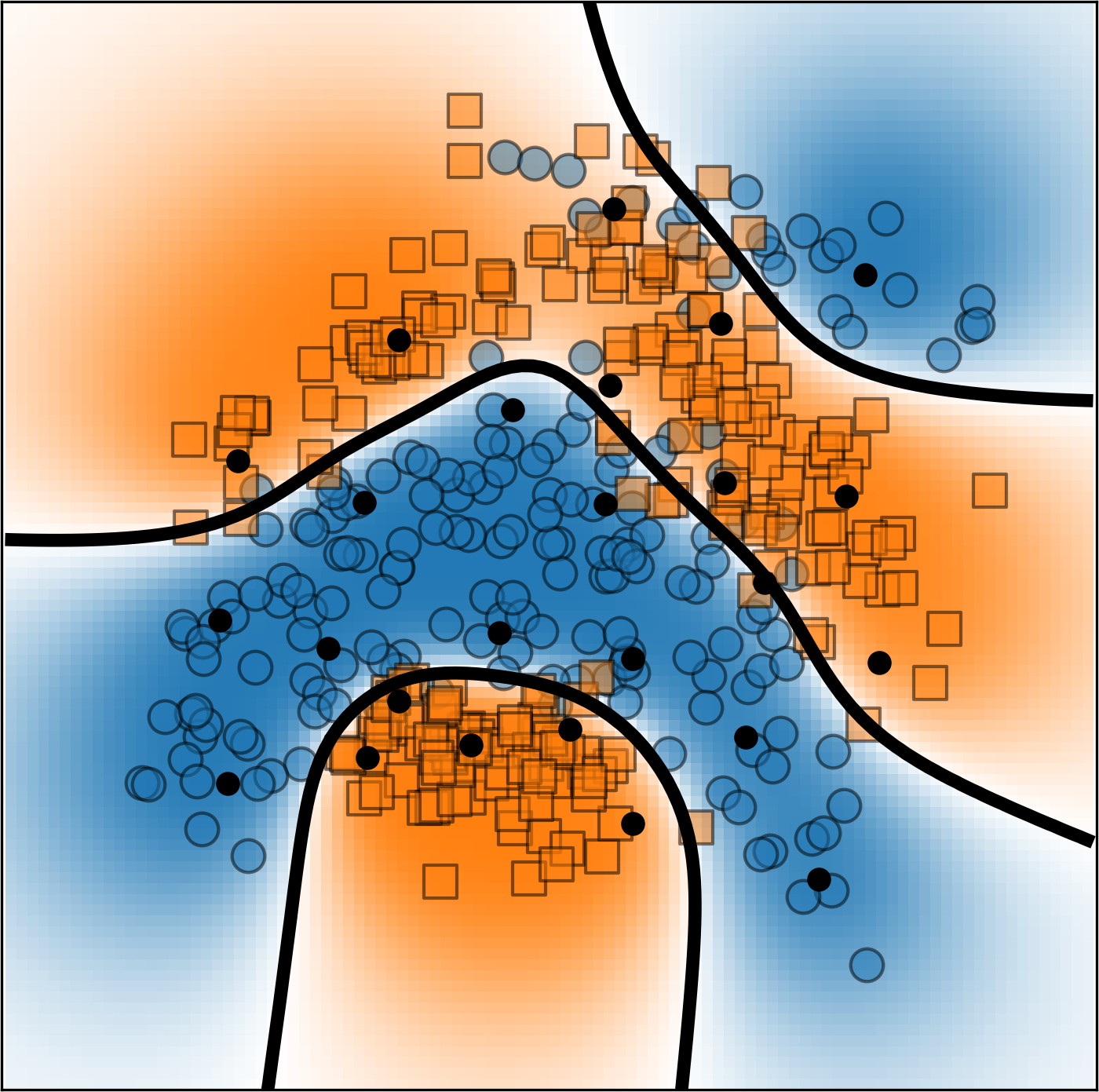}};
		\end{tikzpicture}
		\captionsetup{justification=centering}
		\caption{Offline SVGP model \\ (baseline)}
		\label{fig:app_banana_full_tsvgp}
	\end{subfigure}
	\hfill
	\begin{subfigure}[b]{.3\textwidth}
		\centering
		\setlength{\figurewidth}{.333\textwidth}
		\setlength{\figureheight}{\figurewidth}
		\begin{tikzpicture}[inner sep=0, outer sep=0]
			\foreach \x [count=\i] in {1,2,3} {
				\node (n\i) at (\i\figurewidth,0) {\includegraphics[width=\figurewidth]{fig/banana/appendix/streaming_banana_tsvgp_fc_\x}};
			}
			\setlength{\figurewidth}{\textwidth}
			\setlength{\figureheight}{\figurewidth}
			\node (n4) at (.666\figurewidth,-.666\figurewidth) {\includegraphics[width=\figurewidth]{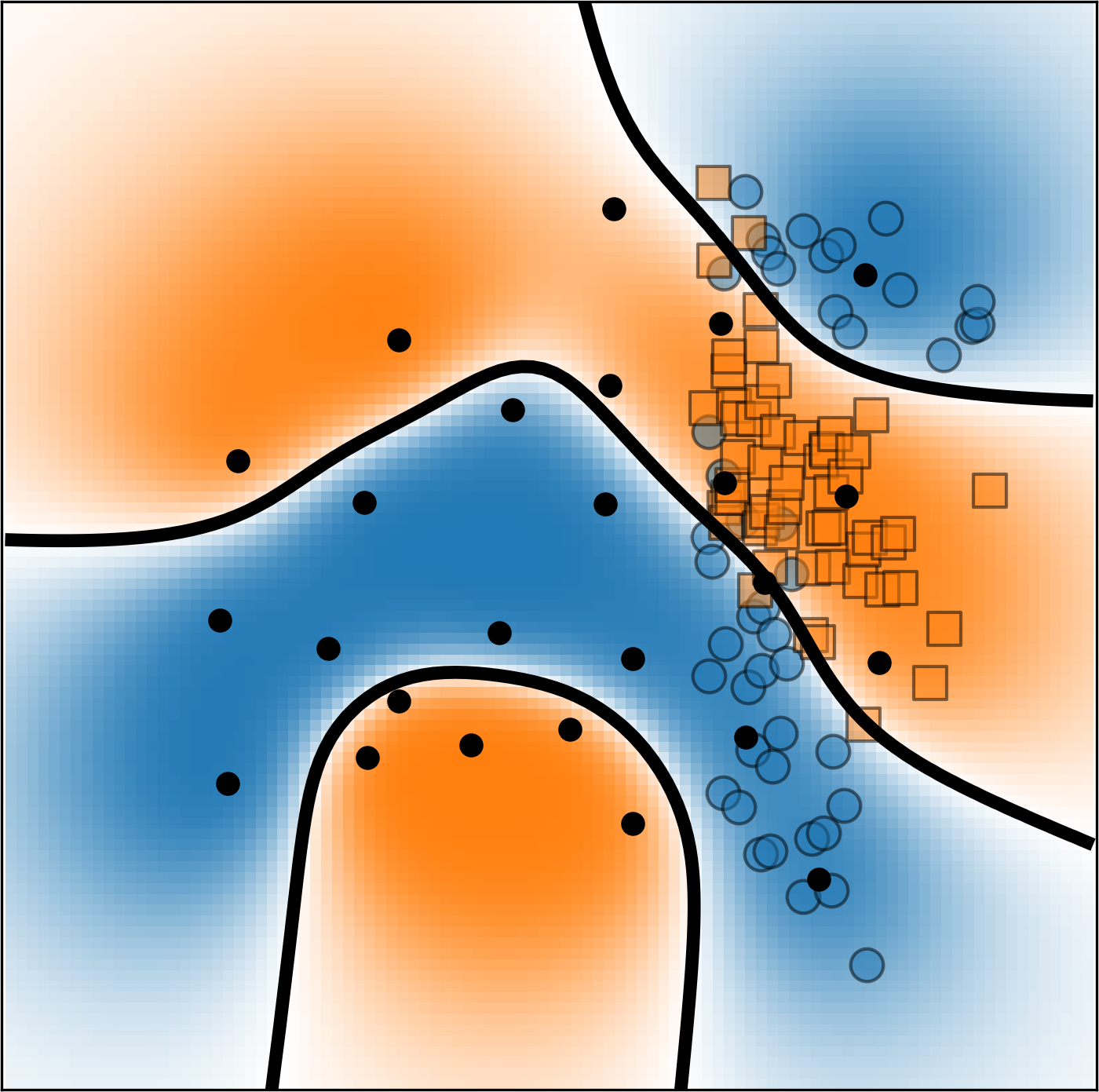}};
			\foreach \i in {1,2,3,4}
			\node[anchor=north west,inner sep=1pt,fill=white,draw=black] at (n\i.north west) {\i};
		\end{tikzpicture}
		\captionsetup{justification=centering}
		\caption{Streaming dual SVGP \\ (ours)}
		\label{fig:app_banana_tsvgp_fc}
	\end{subfigure}
	\hfill
	\begin{subfigure}[b]{.3\textwidth}
		\centering
		\setlength{\figurewidth}{.333\textwidth}
		\setlength{\figureheight}{\figurewidth}
		\begin{tikzpicture}[inner sep=0, outer sep=0]
			\foreach \x [count=\i] in {1,2,3} {
				\node (n\i) at (\i\figurewidth,0) {\includegraphics[width=\figurewidth]{fig/banana/appendix/streaming_banana_online_vargp_\x}};
			}
			\setlength{\figurewidth}{\textwidth}
			\setlength{\figureheight}{\figurewidth}
			\node (n4) at (.666\figurewidth,-.666\figurewidth) {\includegraphics[width=\figurewidth]{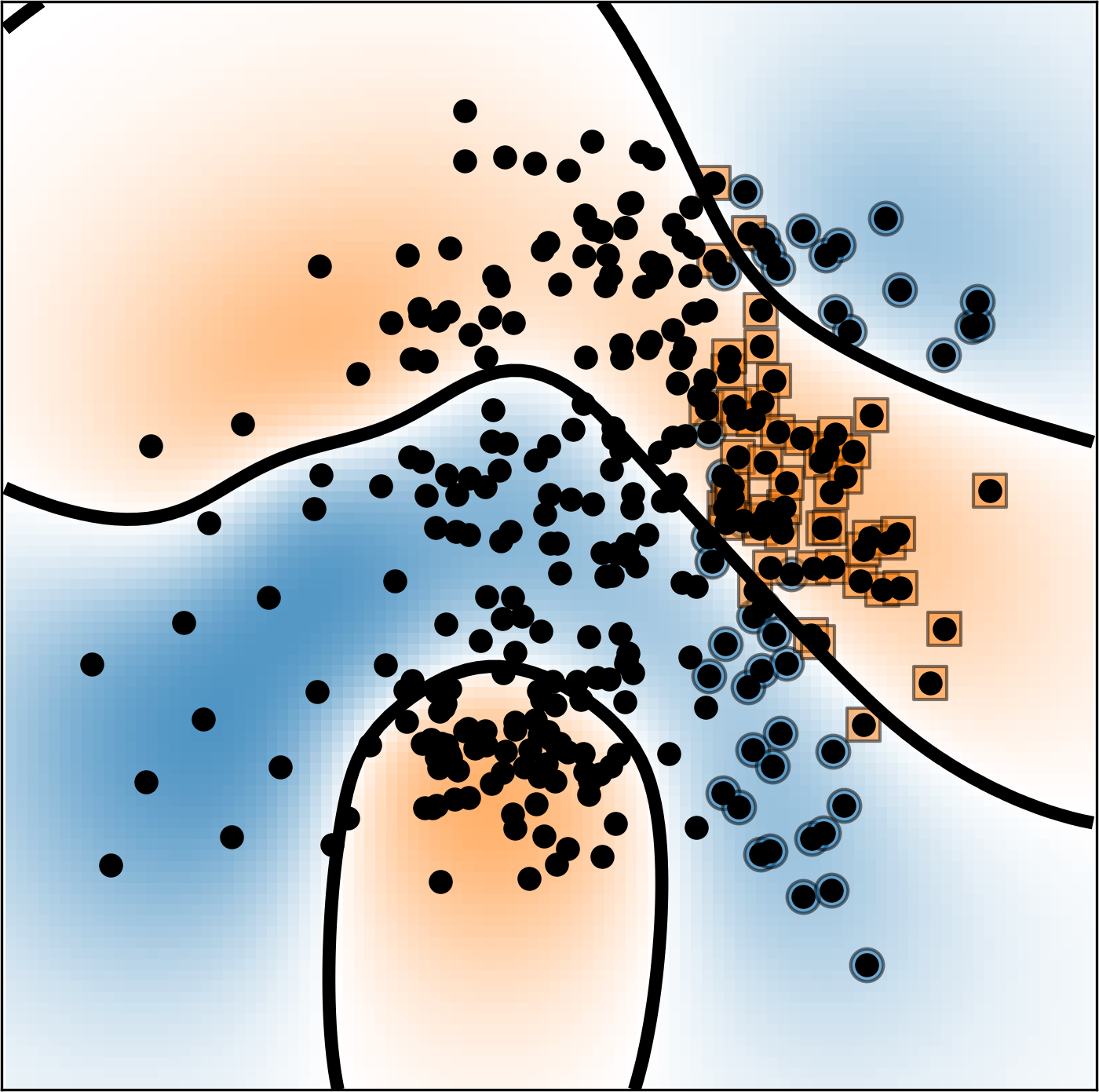}};
			\foreach \i in {1,2,3,4}
			\node[anchor=north west,inner sep=1pt,fill=white,draw=black] at (n\i.north west) {\i};
		\end{tikzpicture}
		\captionsetup{justification=centering}
		\caption{OVC \\ \citep{maddox2021conditioning}}
		\label{fig:app_banana_online_vargp}
	\end{subfigure}
	\caption[Dual conditioning on streaming banana data set]{Conditioning on streaming banana data set; data
		\tikz[baseline=-.6ex, opacity=0.8]{\draw[black, fill=orange] (0, 0) rectangle ++(4pt,4pt);\draw [black, fill=matplotlib-blue] (4pt, 0) circle (2pt);}
		appears batch by batch (1--4).
		The plot shows the decision boundary
		\tikz[baseline=-.6ex,line width=1.5pt]\draw[black](0,0)--(.5,0);
		and the predictive class probability, with colour shading
		\tikz[baseline=1pt]\node[rectangle, anchor=south, right color=orange, left color=white, minimum width=12pt, minimum height=6pt]{};
		and
		\tikz[baseline=1pt]\node[rectangle, anchor=south, right color=matplotlib-blue, left color=white, minimum width=12pt, minimum height=6pt]{};
		increasing the more certain the model is about the class. The inducing points are overlaid as black dots. (a)~Offline SVGP model trained with full data. (b)~Dual SVGP model conditioned on the data appearing in batches. (c)~ Online Variational Conditioning \citep[OVC,][]{maddox2021conditioning} model on batched data.} 
	\label{fig:app_banana_fc}
\end{figure*}

\paragraph{Fast Conditioning}
For a fair comparison against \citet{maddox2021conditioning} in this experiment we also keep the hyperparameters fixed. The problem's challenge is that previous batches are not accessible; only the inferred variational parameters are available. Therefore, online models are needed to condition on new data. As an oracle baseline, we trained an offline SVGP model on the full data (see \cref{fig:app_banana_full_tsvgp}). All three models are intialized with $25$ inducing points and a Mat\'ern-\nicefrac{5}{2} kernel. The hyperparameters for our streaming dual SVGP and \citet{maddox2021conditioning}'s OVC model are taken from the full offline model: only conditioning is performed when the batches of data are received.

We compare decision boundary and predictive class probability of the three models in \cref{fig:app_banana_fc}. The OVC method as introduced by \citet{maddox2021conditioning} essentially initializes new models on each batch and then combines them, hence the increasing number of inducing points for this model. The evolution of the class probability of dual SVGP and OVC is shown in \cref{fig:app_banana_tsvgp_fc} and \cref{fig:app_banana_online_vargp}, respectively. The class probability obtained by the dual SVGP model after seeing the final batch closely matches that of the offline SVGP model. In contrast, the OVC method does not recover the full-data decision boundary, and its uncertainty does not match the offline baseline well. %

\begin{figure*}[t!]
	\centering
	\setlength{\figurewidth}{.3\textwidth}
	\setlength{\figureheight}{\figurewidth}
	\pgfplotsset{scale only axis,axis on top}
	\begin{subfigure}[b]{.3\textwidth}
		\centering
		\begin{tikzpicture}[inner sep=0, outer sep=0]
			\node{\includegraphics[width=\figurewidth]{fig/banana/appendix/streaming_banana_tsvgp_full}};
		\end{tikzpicture}
		\captionsetup{justification=centering}
		\caption{Offline SVGP model}
		\label{fig:app_banana_full_tsvgp_v2}
	\end{subfigure}
	\hfill
	\begin{subfigure}[b]{.3\textwidth}
		\centering
		\setlength{\figurewidth}{.333\textwidth}
		\setlength{\figureheight}{\figurewidth}
		\begin{tikzpicture}[inner sep=0, outer sep=0]
			\foreach \x [count=\i] in {1,2,3} {
				\node (n\i) at (\i\figurewidth,0) {\includegraphics[width=\figurewidth]{fig/banana/appendix/streaming_banana_tsvgp_\x}};
			}
			\setlength{\figurewidth}{\textwidth}
			\setlength{\figureheight}{\figurewidth}
			\node (n4) at (.666\figurewidth,-.666\figurewidth) {\includegraphics[width=\figurewidth]{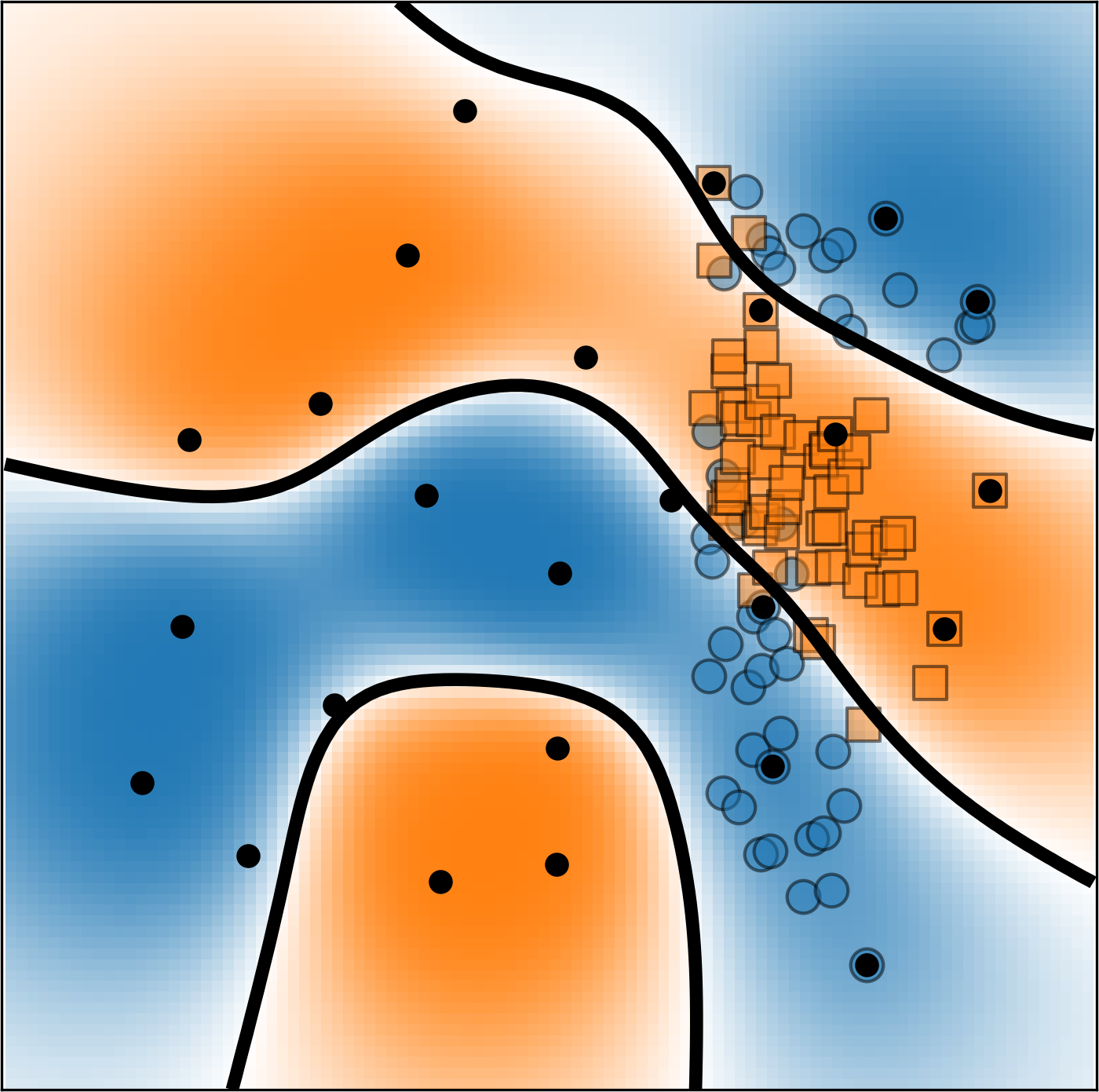}};
			\foreach \i in {1,2,3,4}
			\node[anchor=north west,inner sep=1pt,fill=white,draw=black] at (n\i.north west) {\i};
		\end{tikzpicture}
		\captionsetup{justification=centering}
		\caption{Streaming dual SVGP}
		\label{fig:app_banana_tsvgp}
	\end{subfigure}
	\hfill
	\begin{subfigure}[b]{.3\textwidth}
		\centering
		\setlength{\figurewidth}{.333\textwidth}
		\setlength{\figureheight}{\figurewidth}
		\begin{tikzpicture}[inner sep=0, outer sep=0]
			\foreach \x [count=\i] in {1,2,3} {
				\node (n\i) at (\i\figurewidth,0) {\includegraphics[width=\figurewidth]{fig/banana/appendix/streaming_banana_bui_\x}};
			}
			\setlength{\figurewidth}{\textwidth}
			\setlength{\figureheight}{\figurewidth}
			\node (n4) at (.666\figurewidth,-.666\figurewidth) {\includegraphics[width=\figurewidth]{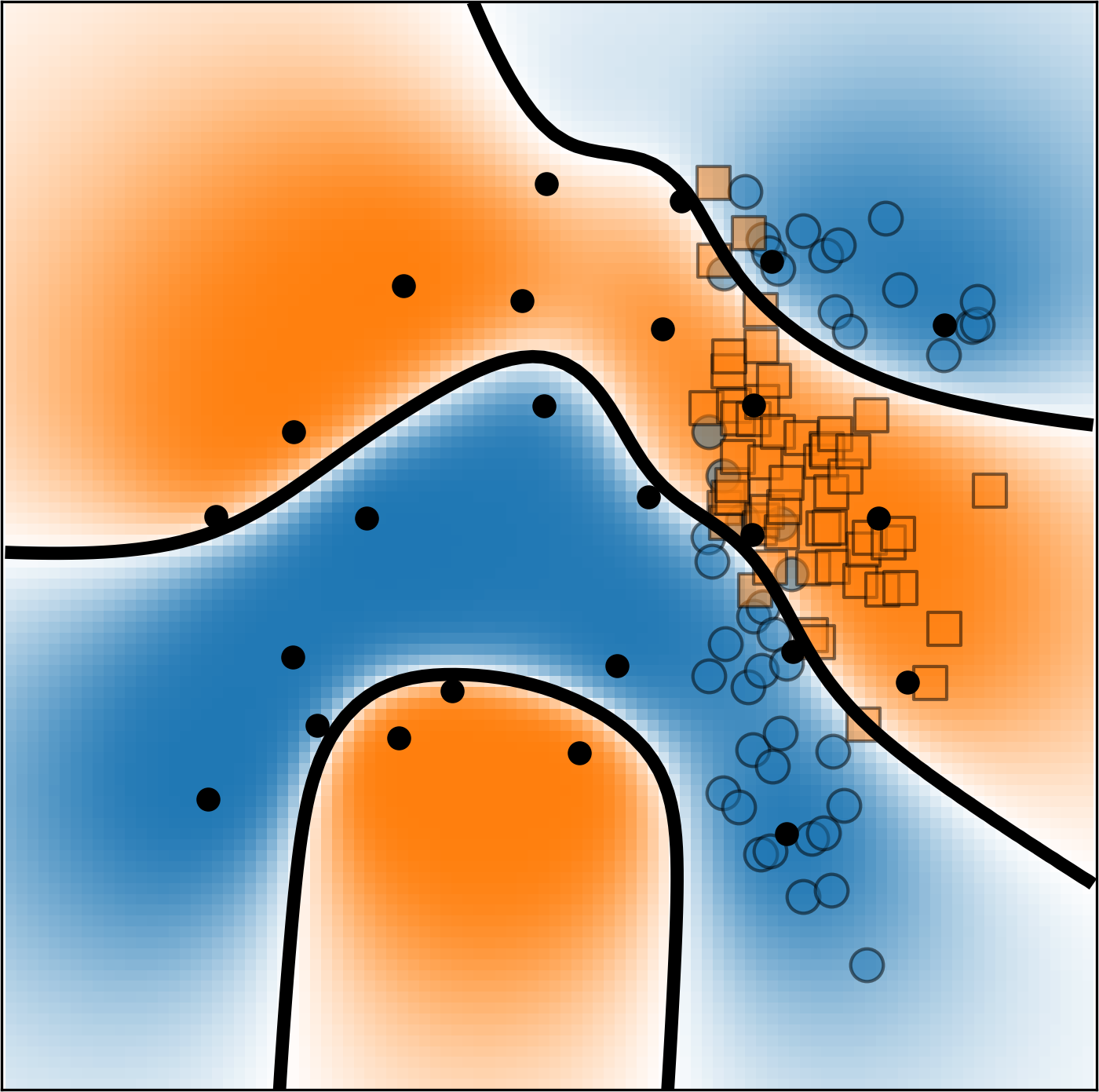}};
			\foreach \i in {1,2,3,4}
			\node[anchor=north west,inner sep=1pt,fill=white,draw=black] at (n\i.north west) {\i};
		\end{tikzpicture}
		\captionsetup{justification=centering}
		\caption{\citet{bui2017streaming}}
		\label{fig:app_banana_bui}
	\end{subfigure}
		\caption[Streaming banana data set]{Streaming banana data set when  $(\valpha_{\vu},\vBeta_{\vu},\vtheta,\MZ)$ are learnt; data
		\tikz[baseline=-.6ex, opacity=0.8]{\draw[black, fill=orange] (0, 0) rectangle ++(4pt,4pt);\draw [black, fill=matplotlib-blue] (4pt, 0) circle (2pt);}
		appears batch by batch (1--4).
		The plot shows the decision boundary
		\tikz[baseline=-.6ex,line width=1.5pt]\draw[black](0,0)--(.5,0);
		and the predictive class probability, with colour shading
		\tikz[baseline=1pt]\node[rectangle, anchor=south, right color=orange, left color=white, minimum width=12pt, minimum height=6pt]{};
		and
		\tikz[baseline=1pt]\node[rectangle, anchor=south, right color=matplotlib-blue, left color=white, minimum width=12pt, minimum height=6pt]{};
		increasing the more certain the model is about the class. The inducing points are overlaid as black dots.}
	\label{fig:app_banana_learn}
\end{figure*}

\paragraph{Hyperparameter Learning}
Next, we conduct an experiment in which we learn all the hyperparameters $(\valpha_{\vu},\vBeta_{\vu},\vtheta,\MZ)$, in contrast to the previous experiment where the hyperparameters were fixed.
We compare to \citet{bui2017streaming}, who previously considered a similar test setup. Again, as an oracle baseline, an offline SVGP model is trained on the full data (see \cref{fig:app_banana_full_tsvgp_v2}). All three models are intialized with $25$ inducing points and Mat\'ern-\nicefrac{5}{2} kernel. The hyperparameters for the dual SVGP are optimized using Adam optimizer \citep{adam} with learning rate $10^{-2}$ and for \citet{bui2017streaming} we use L-BFGS. (for \citet{bui2017streaming}'s model, we tried both Adam optimizer and L-BFGS optimizer; L-BFGS gave better results).

We compare the decision boundary and predictive class probability of the three models in \cref{fig:app_banana_learn}. The evolution of the class probability of our dual SVGP and \citet{bui2017streaming} is shown in \cref{fig:app_banana_tsvgp} and \cref{fig:app_banana_bui}.

\subsection{Split MNIST}
\label{appendix:mnist}
For split MNIST (see \cref{sec:mnist}), we use the standard MNIST data provided by TensorFlow. We concatenate the standard train and test set provided and split it $80:20$ for training and testing. Each task is sub-divided into batches of $4000$.

Both the models, our proposed model and the \citeauthor{bui2017streaming} model, use a Mat\'ern-\nicefrac{5}{2} kernel initialized with unit variance and lengthscale. We use 10 latent GPs which matches the number of classes, with $300$ inducing variables, and use a softmax likelihood.

For hyperparameter learning in our proposed model, we use the Adam optimizer \citep{adam} with learning rate $10^{-2}$ for $100$ iterations for each set of data. The number of memory points for each set of tasks is set to $400$. We also found that for this task the removal of memory set from the variational parameters actually gave worse performance for split MNIST so do not perform the removal the last term of \cref{eq:elbo_new}, we suspect this is due to the highly non-stationary nature of the problem and how double counting dual variables alleviates the difficulty. 

For \citeauthor{bui2017streaming}, we experiment with the L-BFGS optimizer as well as Adam optimizer with $50$ and $100$ iterations; in this case we found that Adam works better (L-BFGS fails to learn subsequent tasks). The accuracy over tasks for different numbers of iterations can be seen in \cref{tbl:split_mnist_acc_bui,tbl:split_mnist_acc_bui_v2}.

	\begin{table}[b!]
		\begin{minipage}[t][4.75cm][t]{.45\textwidth}
			\centering\footnotesize  
			\caption{Test accuracy (and standard deviation over different random seeds) on {\em split MNIST} over all tasks thus far for \citet{bui2017streaming}. Low accuracy over all previous tasks shows that the method suffers from forgetting. (Adam optimizer, 50 iteration loops)}
			\label{tbl:split_mnist_acc_bui}
			\setlength{\tabcolsep}{3pt}
			\begin{tabularx}{\textwidth}{c c c c c c} 
				\toprule
				\textbf{Task} & \digit{0}, \digit{1} & \digit{2}, \digit{3} & \digit{4}, \digit{5} & \digit{6}, \digit{7} & \digit{8}, \digit{9}\\
				\midrule
				\#1	   &$.72 (.01)$  & &  &  &  \\
				\#2   &  $.11 (.01)$ & $.36 (.26)$  &  &  & \\
				\#3   &  $.04 (.02)$ & $.02 (.00)$  & $.95 (.01)$  &  & \\
				\#4	  &  $.02 (.01)$ & $.01 (.00)$  & $.03 (.01)$ & $.98 (.00)$ & \\
				\#5	  &  $.06 (.04)$ & $.06 (.04)$  & $.05 (.04)$ & $.10 (.04)$ & $.55 (.34)$\\
				\bottomrule
			\end{tabularx}
		\end{minipage}
	\hfill
	\begin{minipage}[t][4.75cm][t]{.45\textwidth}
		\centering\footnotesize  
		\caption{Test accuracy (and standard deviation over different random seeds) on {\em split MNIST} over all tasks thus far for \citet{bui2017streaming}. Low accuracy over all previous tasks shows that the method suffers from forgetting. (Adam optimizer, 100 iteration loops)}
		\label{tbl:split_mnist_acc_bui_v2}
		\setlength{\tabcolsep}{3pt}
		\begin{tabularx}{\textwidth}{c c c c c c} 
			\toprule
			\textbf{Task} & \digit{0}, \digit{1} & \digit{2}, \digit{3} & \digit{4}, \digit{5} & \digit{6}, \digit{7} & \digit{8}, \digit{9}\\
			\midrule
			\#1	   &$.99 (.00)$  & &  &  &  \\
			\#2   &  $.01 (.00)$ & $.96 (.00)$  &  &  & \\
			\#3   &  $.02 (.04)$ & $.02 (.04)$  & $.80 (.40)$  &  & \\
			\#4	  &  $.00 (.00)$ & $.00 (.00)$  & $.00 (.00)$ & $.99 (.00)$ & \\
			\#5	  &  $.00 (.00)$ & $.00 (.00)$  & $.00 (.00)$ & $.00 (.00)$ & $.97 (.00)$\\
			\bottomrule
		\end{tabularx}
	\end{minipage}

	\end{table}

\subsection{Magnetic Anomaly Modelling}
\label{appendix:magnetic}
For the robot experiment for learning magentic field anomalies, we use the data from \citet{solin2018modeling} that is available at \url{https://github.com/AaltoML/magnetic-data}. The exact metric position of the robot is provided time-synced to magnetometer samples from a low-cost/high-noise Invensense MPU-9150 magnetometer sampled at 50~Hz. As discussed in \citet{solin2018modeling}, GPs provide a principled approach for modelling smooth anomalies in the magnetic vector field.

We consider the task of online mapping of local anomalies in the ambient Earth magnetic field. These anomalies are induced by the bedrock and magnetic material in the building structure indoors. We follow the experiment setup and data provided in \citet{solin2018modeling}, where a small wheeled robot equipped with a 3-axis magnetometer moves around in an indoor space measuring roughly $\SI{6}{\meter} \times \SI{6}{\meter}$. We consider a simplified model, where we assign a GP prior to the magnetic field strength $\|\MH\| \sim \mathcal{GP}(0,\sigma_0^2+\kappa^\mathrm{Mat.}_{\sigma^2,\ell}(\vx,\vx'))$ (in \si{\micro\tesla}) over the space under presence of Gaussian measurement noise with variance $\sigma_\mathrm{n}^2$.

We include two separate experiments with different data paths in the same space. The first experiment (\cref{fig:robot-a}) uses four separate paths of the robot that cover slightly different parts of the space. Under the sequential learning framework, we simultaneously learn the four hyperparameters ($\sigma_0^2, \sigma^2, \ell, \sigma_\mathrm{n}^2$) and the representation in terms of inducing points and memory. The high measurement noise renders the value of single data points small and stresses the importance of the sparse approach. The problem is sequential, as we only receive information as the robot moves through the input space and do not have access to all previous data.  Our sequential method is able to form a representation by spreading inducing points and learning the hyperparameters progressively. Showing the method has practical importance in real-world robot estimation tasks, where the robot is not constrained to a predefined area. This has been a weakness in previous approaches that have considered a fixed domain to form an efficient sparse basis function decomposition of the problem in that domain. In \cref{fig:robot-b}, the set up of the second experiment is similar, but now we receive data during each robot path instead after one exploration. The visualization shows the mean estimate with marginal variance (uncertainty) controlling the opacity. We recover the same local estimate as in experiment \cref{fig:robot-a}.

The data is a set of $9$ trajectories out of which we use $5$ for the experiments. For Experiment~\#1, \cref{fig:robot-a}, we use trajectory 1, 2, 4, and 5 (with $n=8875,9105,7332,8313$, respectively) and for Experiment~\#2, \cref{fig:robot-b}, we use trajectory 3 ($n=9404$). For both the experiments, we use a sum of two kernels: constant kernel and a Mat\'ern-\nicefrac{5}{2} kernel. For the constant kernel, we set the initial variance as $500$. The number of inducing variables is set to $100$ in both the experiments, and we use a Gaussian likelihood initialized with a noise variance of $0.1$. For optimization of the hyperparameters, we use the Adam optimizer with learning rate $10^{-2}$ for $20{,}000$ iterations. For Experiment~\#2, we split trajectory~3 into 20 sets, thus each set has around $470$ data points. 

\subsection{Lunar Landing}
\label{appendix:lunar}
For the lunar landing experiment (see \cref{sec:BO_AL}), we use a combination of two models: a regression model with the aim to increase the reward and a classification model for success or failure. Both the models are the proposed dual SVGP model with Gaussian likelihood and Bernoulli likelihood, respectively. For the regression model, we use an ARD Mat\'ern-\nicefrac{5}{2} kernel with initial variance calculated from the initial observation data and the lengthscale initialized with $0.2$. We use a Gaussian likelihood with an initial unit noise variance. For the classification model, we use an ARD squared exponential kernel with the magnitude initialized with $100$ and lengthscale with $0.2$. For the classification model, we fix the magnitude. Both models use $100$ inducing points. The acquisition function is the product of ExpectedImprovement from the regression model and the ProbabilityOfValidity from the classification model. Initial data for both the batch version and the non-batch version is the same set of $24$ points. For the batch models, we use a batch-size of 3 and both the models are optimized for $90$ function evaluations. We run the experiment with 5 random initial observations and plot the mean and individual rewards along with the BO iterations in \cref{fig:lunar_landing}.

\subsection{UCI Data Sets}
\label{appendix:uci}
We benchmark on UCI data sets both for regression and classification tasks (see \cref{sec:streaming}). We report negative log predictive density (NLPD), root mean square error (RMSE) for regression tasks, and classification error for classification tasks in \cref{tbl:uci_experiment_full} and \cref{tbl:uci_experiment_eval} over 10-fold cross-validation. We run three models: \textbf{offline} model and two online models (\textbf{Ours}, and  \textbf{\citeauthor{bui2017streaming}}).  The offline model has access to the whole data and is used as baseline. \textbf{Ours} is our proposed method where all the parameters $(\valpha_{\vu},\vBeta_{\vu}, \vtheta, \MZ)$ are learnt. \textbf{\citeauthor{bui2017streaming}} is the model proposed by \citet{bui2017streaming}.

All the models use a Mat\'ern-\nicefrac{5}{2} kernel with lengthscale and variance initialized to $1.0$, a Gaussian likelihood with initial noise variance $0.1$, and $100$ inducing points. For converting the data sets into a streaming setting, we sort the data on the first dimension and split the data set into $50$ subsets for all data sets apart from \textit{Mammographic} (20 subsets). For variational parameters $(\valpha_{\vu},\vBeta_{\vu})$ of our proposed model, we use natural gradient updates, with learning rate 0.8 and 2 update steps for regression and learning rate 0.5 and 4 update steps for classification. For learning hyperparameters $(\theta, \MZ)$, we use the Adam optimizer \citep{adam} with learning rate $10^{-2}$ and $100$ update steps. For optimizing the parameters of \citeauthor{bui2017streaming}'s model, we tried both Adam and second-order optimization using L-BFGS. In our experiments, we found that L-BFGS  with 100 iteration steps or until the default convergence condition is met works the best, and used this for the final results. The offline model is trained using Adam optimizer with learning rate $10^{-2}$.

\begin{table}[h]
    \centering
	\begin{minipage}[t][4.75cm][t]{.75\textwidth}
		\raggedright
		\caption{Root mean square error (RMSE) (for regression tasks,~$^R$) and classification error (for classification tasks,~$^C$) on 10-fold cross-validation for UCI data sets, lower is better. }
		\label{tbl:uci_experiment_eval}
		\footnotesize
		\setlength{\tabcolsep}{8pt}
		\centering
		\begin{tabularx}{\textwidth}{l l l l l} 
			\toprule
			\textbf{Data set} & \textbf{Dimension ($N$, $D$)}  & \textbf{Offline} &  \textbf{Ours} & \textbf{\citeauthor{bui2017streaming}} \\
			\midrule
			Elevators$^R$ 			    & $(16599, 18)$	  &  $.39 (.00)$ & $.42 (.01)$ & $.42 (.01)$\\
			Bike$^R$		 			   & $(17379, 17)$	  &  $.29 (.01)$  & $.37 (.01)$  & $.38 (.01	)$ \\
			Mammographic$^C$   & $(961, 6)$ 		 & $.18 (.01)$ & $.81 (.03)$  & $.81 (.04)$ \\
			Bank$^C$					 &	$(4521, 17)$     & $.11 (.01)$ & $.88 (.01)$      & $.89 (.02)$ \\
			Mushroom$^C$  	  	   &  $(8124, 22)$     & $.00 (.00)$ & $.97 (.03)$		& $.99 (.01)$\\
			Adult$^C$ 					 &	$(48842, 15)$   &$.16 (.00)$	& $.82 (.01)$  	& $.83 (.01)$ \\
			\bottomrule
		\end{tabularx}
	\end{minipage}
\end{table}

\subsection{UCI Data sets: BLS and Random}
\label{appendix:uci_random_bls}
We experiment with two different techniques for updating memory: random and the proposed Bayesian leverage score (BLS). We report the negative log predictive density (NLPD) (\cref{tbl:uci_experiment_bls_random}) and showcase BLS outperforming random memory technique.

All the models use a Mat\'ern-\nicefrac{5}{2} kernel with lengthscale and variance initialized to $1.0$, a Gaussian likelihood with initial noise variance $0.1$, and $100$ inducing points. For converting the data sets into a streaming setting, we sort the data on the first dimension and split the data set into $10$ subsets for all data sets. For variational parameters $(\valpha_{\vu},\vBeta_{\vu})$ of our proposed model, we use natural gradient updates, with learning rate 0.8 and 2 update steps for regression and learning rate 0.2 and 10 update steps for classification. For learning hyperparameters $(\theta, \MZ)$, we use the Adam optimizer \citep{adam} with learning rate $10^{-2}$ and $100$ update steps. For optimizing the parameters of \citeauthor{bui2017streaming}'s model, we tried both Adam and second-order optimization using L-BFGS. In our experiments, we found that L-BFGS  with 100 iteration steps or until the default convergence condition is met works the best, and used this for the final results. The offline model is trained using Adam optimizer with learning rate $10^{-2}$.

\begin{table}[h]
	\centering\footnotesize
		\begin{minipage}[t][4.75cm][t]{.38\textwidth}
	\caption{Negative log predictive density (NLPD) on 5-fold cross-validation for UCI data sets, lower is better, for both random and Bayesian leverage score.}
		\label{tbl:uci_experiment_bls_random}
		\setlength{\tabcolsep}{6pt}
		\begin{tabularx}{\columnwidth}{l l l} 
			\toprule
			\textbf{Data set} & \textbf{Random} & \textbf{BLS} \\
			\midrule
			Adult					 & $.35 (.01)$  & $.34 (.01)$ \\
			Bank 					& $.29 (.03)$ & $.27 (.03)$\\
			Mushroom		  & $.03 (.00)$ & $.03 (.00)$\\
			Mammographic & $.45 (.03)$ & $.43 (.02)$\\
			Elevators 			& $.63 (.02) $& $.63(.04)$\\
			Bike 					& $.47 (.05)$ &  $.46 (.03)$ \\
			\bottomrule
		\end{tabularx}
	
	\end{minipage}
	\end{table}

\end{document}